\documentclass[journal,compsoc]{IEEEtran}
\usepackage{amssymb}
\usepackage{amsmath}
\usepackage{amsfonts}
\usepackage{times}
\usepackage{subfigure}
\usepackage{mathtools}
\usepackage{graphicx}
\usepackage{setspace}
\usepackage{multirow}
\usepackage{hyperref}
\usepackage{color}
\usepackage[switch]{lineno}
\usepackage{url}
\usepackage{bm}
\usepackage{array}
\usepackage[top=1in, bottom=1in, left=1.25in, right=1.25in]{geometry}

\newcolumntype{P}[1]{>{\raggedright\arraybackslash}p{#1}}
\newcolumntype{M}[1]{>{\raggedright\arraybackslash}m{#1}}
\DeclareMathOperator*{\argmin}{arg\,min}
\DeclareMathOperator*{\argmax}{arg\,max}

\DeclarePairedDelimiter\abs{\lvert}{\rvert}%
\DeclarePairedDelimiter\norm{\lVert}{\rVert}%

\ifCLASSOPTIONcompsoc
  \usepackage[nocompress]{cite}
\else
  \usepackage{cite}
\fi
\ifCLASSINFOpdf
\else
\fi
\begin{document}
%
\title{Machine Learning for Spatiotemporal Sequence Forecasting: A Survey}
%
%
%
%

\author{Xingjian~Shi,
        Dit-Yan~Yeung,~\IEEEmembership{Senior Member,~IEEE}
\IEEEcompsocitemizethanks{\IEEEcompsocthanksitem Xingjian Shi is with the Department of Computer Science and Engineering, the Hong Kong University of Science and Technology. E-mail: \href{mailto:xshiab@cse.ust.hk}{xshiab@cse.ust.hk}
\IEEEcompsocthanksitem Dit-Yan Yeung is with the Department of Computer Science and Engineering, the Hong Kong University of Science and Technology. E-mail: \href{mailto:dyyeung@cse.ust.hk}{dyyeung@cse.ust.hk}}
\thanks{}}

\IEEEtitleabstractindextext{%
\begin{abstract}
Spatiotemporal systems are common in the real-world. Forecasting the multi-step future of these spatiotemporal systems based on the past observations, or, \emph{Spatiotemporal Sequence Forecasting} (STSF), is a significant and challenging problem. Although lots of real-world problems can be viewed as STSF and many research works have proposed machine learning based methods for them, no existing work has summarized and compared these methods from a unified perspective. This survey aims to provide a systematic review of machine learning for STSF. In this survey, we define the STSF problem and classify it into three subcategories: \emph{Trajectory Forecasting of Moving Point Cloud} (TF-MPC), \emph{STSF on Regular Grid} (STSF-RG) and \emph{STSF on Irregular Grid} (STSF-IG). We then introduce the two major challenges of STSF: 1) how to learn a model for multi-step forecasting and 2) how to adequately model the spatial and temporal structures. After that, we review the existing works for solving these challenges, including the general learning strategies for multi-step forecasting, the classical machine learning based methods for STSF, and the deep learning based methods for STSF. We also compare these methods and point out some potential research directions.
\end{abstract}

}

\maketitle

\IEEEdisplaynontitleabstractindextext

%
\IEEEpeerreviewmaketitle

\IEEEraisesectionheading{\section{Introduction}\label{ch:intro}}

%
%
%
%
\IEEEPARstart{M}{any} real-world phenomena are spatiotemporal, such as the traffic flow, the diffusion of air pollutants and the regional rainfall. Correctly predicting the future of these spatiotemporal systems based on the past observations is essential for a wide range of scientific studies and real-life applications like traffic management, precipitation nowcasting, and typhoon alert systems.

If there exists an accurate numerical model of the dynamical system, which usually happen in the case that we have fully understood its rules, forecasting can be achieved by first finding the initial condition of the model and then simulate it to get the future predictions~\cite{hunt2007efficient}. However, for some complex spatiotemporal dynamical systems like atmosphere, crowd, and natural videos, we are still not entirely clear about their inner mechanisms. In these situations, machine learning based methods, by which we can try to learn the systems' inner ``laws'' based on the historical data, have proven helpful for making accurate predictions~\cite{ranzato2014video,zheng2015forecasting,alahi2016social,xingjian2015convolutional,mathieu2016deep}. Moreover, recent studies~\cite{nitish2015unsupervised,finn2017deep} have shown that the ``laws'' inferred by machine learning algorithms can further be used to guide other tasks like classification and control. For example, controlling a robotic arm will be much easier once you can anticipate how it will move after performing specific actions~\cite{finn2017deep}.

\begin{table*}[tb!]
  \centering
  \caption{Types of spatiotemporal sequence forecasting problems.}
  \label{tbl:type-spatiotemporal-forecasting}
  \begin{tabular}{|l|l|l|}
    \hline
    Problem Name                                    & Coordinates       & Measurements \\ \hline
    Trajectory Forecasting of Moving Point Cloud                    & Changing                      & Fixed/Changing \\ \hline
    Spatiotemporal Forecasting on Regular Grid & Fixed regular grid & Changing \\ \hline
    Spatiotemporal Forecasting on Irregular Grid & Fixed irregular grid & Changing \\ \hline
  \end{tabular}
\end{table*}

In this paper, we review these machine learning based methods for spatiotemporal sequence forecasting.
We define a length-$T$ spatiotemporal sequence as a series of matrices $\mathbf{X}_{1:T} = [\mathbf{X}_1, \mathbf{X}_2, ... \mathbf{X}_T]$. Each $\mathbf{X}_t \in \mathbb{R}^{K \times (D+E)}$ contains a set of coordinates and their corresponding measurements. Here, $K$ is the number of coordinates, $D$ is the number of measurements, and $E$ is the dimension of the coordinate. We further denote the measurement part and coordinate part of $\mathbf{X}_t$ as $\mathbf{M}_t \in \mathbb{R}^{K \times D}$ and $\mathbf{C}_t \in \mathbb{R}^{K \times E}$. For many forecasting problems, we can use auxiliary information to enhance the prediction accuracy. For example, regarding wind speed forecasting, the latitudes and past temperatures are also helpful for predicting the future wind speeds. We denote the auxiliary information available at timestamp $t$ as $\mathcal{A}_t$. The \emph{Spatiotemporal Sequence Forecasting} (STSF) problem is to predict the length-$L$ ($L>1$) sequence in the future given the previous observations plus the auxiliary information that is allowed to be empty. The mathematical form is given in~\eqref{eq:stsf}.
\begin{equation}
\label{eq:stsf}
\begin{aligned}
	\mathbf{\hat{X}}_{t+1:t+L} = \argmax_{\mathbf{X}_{t+1:t+L}} p(\mathbf{X}_{t+1:t+L} \mid \mathbf{X}_{1:t}, \mathcal{A}_{t}).
\end{aligned}
\end{equation}
Here, we limit STSF to only contain problems where the input and output are both sequences with multiple elements. According to this definition, problems like video generation using a static image~\cite{vondrick2016generating} and dense optical flow prediction~\cite{walker2015dense}, in which the input or the output contain a single element, will not be counted.

Based on the characteristics of the coordinate sequence $\mathbf{C}_{1:T}$ and the measurement sequence $\mathbf{M}_{1:T}$, we classify STSF into three categories as shown in Table~\ref{tbl:type-spatiotemporal-forecasting}.
If the coordinates are changing, the STSF problem is called \emph{Trajectory Forecasting of Moving Point Cloud} (TF-MPC). Problems like human trajectory prediction~\cite{yamaguchi2011you, alahi2016social} and human dynamics prediction~\cite{taylor2006modeling, fragkiadaki2015recurrent, jain2016structural} fall into this category. We need to emphasize here that the entities in moving point cloud can not only be human but also be general moving objects. If the coordinates are fixed, we only need to predict the future measurements. Based on whether the coordinates lie in a regular grid, we can get the other two categories, called \emph{STSF on Regular Grid} (STSF-RG) and \emph{STSF on Irregular Grid} (STSF-IG). Problems like precipitation nowcasting~\cite{xingjian2015convolutional}, crowd density prediction~\cite{ali2018periodic} and video prediction~\cite{ranzato2014video}, in which we have dense observations, are STSF-RG problems. Problems like air quality forecasting~\cite{zheng2015forecasting}, influenza prediction~\cite{senanayake2016predicting} and traffic speed forecasting~\cite{yu2017deep}, in which the monitoring stations (or sensors) spread sparsely across the city, are STSF-IG problems. We need to emphasize that although the measurements in TF-MPC can also be time-variant due to factors like appearance change, most models for TF-MPC deal with the fixed-measurement case and only predict the coordinates. We thus focus on the fixed-measurement case in this survey. 

Having both the characteristics of spatial data like images and temporal data like sentences and audios, spatiotemporal sequences contain information about ``what'', ``when'', and ``where'' and provide a comprehensive view of the underlying dynamical system. However, due to the complex spatial and temporal relationships within the data and the potential long forecasting horizon, spatiotemporal sequence forecasting imposes new challenges to the machine learning community. The first challenge is how to learn a model for multi-step forecasting. Early researches in time-series forecasting~\cite{cox1961prediction, chevillon2007direct} have investigated two approaches: \emph{Direct Multi-step} (DMS) estimation and \emph{Iterated Multi-step} (IMS) estimation. The DMS approach directly optimizes the multi-step forecasting objective while the IMS approach learns a one-step-ahead forecaster and iteratively applies it to generate multi-step predictions. However, choosing between DMS and IMS involves a trade-off among forecasting bias, estimation variance, the length of the prediction horizon and the model's nonlinearity~\cite{taieb2014boosting}. Recent studies have tried to find the mid-ground between these two approaches~\cite{taieb2014boosting,bengio2015scheduled}. The second challenge is how to model the spatial and temporal structures within the data adequately. In fact, the number of free parameters of a length-$T$ spatiotemporal sequence can be up to $O(K^T D^T)$ where $K$ is usually above thousands for STSF-RG problems like video prediction. Forecasting in such a high dimensional space is impossible if the model cannot well capture the data's underlying structure. Therefore, machine learning based methods for STSF, including both classical methods and deep learning based methods, have special designs in their model architectures for capturing these spatiotemporal correlations.

In this paper, we organize the literature about STSF following these two challenges. Section~\ref{ch:multistep} covers the general learning strategies for multi-step forecasting including IMS, DMS, boosting strategy and scheduled sampling. Section~\ref{ch:method-classical} reviews the classical methods for STSF including the feature-based methods, state-space models, and Gaussian process models. Section~\ref{ch:dl} reviews the deep learning based methods for STSF including deep temporal generative models and feed-forward and recurrent neural network-based methods. Section~\ref{ch:discussion} summarizes the survey and discusses some future research directions.

\section{Learning Strategies for Multi-step Forecasting}
\label{ch:multistep}
Compared with single-step forecasting, learning a model for multi-step forecasting is more challenging because the forecasting output $\mathbf{\hat{X}}_{t+1:t+L}$ is a sequence with non-i.i.d elements.
In this section, we first introduce and compare two basic strategies called IMS and DMS~\cite{chevillon2007direct} and then review two extensions that bridge the gap between these two strategies. To simplify the notation, we keep the following notation rules in this and the following sections. We denote the information available at timestamp $t$ as $\mathcal{F}_t = \{\mathbf{X}_{1:t}, \mathcal{A}_{t}\}$. Learning an STSF model is thus to train a model, which can be either probabilistic or deterministic, to make the $L$-step-ahead prediction $\mathbf{\hat{X}}_{t+1:t+L}$ based on $\mathcal{F}_t$. We also denote the ground-truth at timestamp $t$ as $\mathbf{\tilde{X}}_t$ and recursively applying the basic function $f(x)$ for $h$ times as $f^{(h)}(x)$. Setting the $k$th subscript to be ``:'' means to select all elements at the $k$th dimension. Setting the $k$th subscript to a letter $i$ means to select the $i$th element at the $k$th dimension. For example, for matrix $\mathbf{X}$, $\mathbf{X}_{i, :}$ means the $i$th row, $\mathbf{X}_{:, j}$ means the $j$th column, and $\mathbf{X}_{i, j}$ means the $(i, j)$th element. The flattened version of matrices $\mathbf{A}$, $\mathbf{B}$ and $\mathbf{C}$ is denoted as $\text{vec}(\mathbf{A}, \mathbf{B}, \mathbf{C})$.
\subsection{Iterative Multi-step Estimation}
The IMS strategy trains a one-step-ahead forecasting model and iteratively feeds the generated samples to the one-step-ahead forecaster to get the multi-step-ahead prediction. There are two types of IMS models, the deterministic one-step-ahead forecasting model $\mathbf{X}_{t} = m(\mathcal{F}_t;\phi)$ and the probabilistic one-step-ahead forecasting model $p(\mathbf{X}_{t} \mid \mathcal{F}_t;\phi)$. For the deterministic forecasting model,
the forecasting process and the training objective of the deterministic forecaster~\cite{taieb2014boosting} are shown in~\eqref{eq:ims-testing} and~\eqref{eq:ims-training-deterministic} respectively where $d(\tilde{x}, x)$ measures the distance between $\tilde{x}$ and $x$.
\begin{equation}
	\label{eq:ims-testing}
	\begin{aligned}
		\mathbf{\hat{X}}_{t+h} &= m^{(h)}(\mathcal{F}_t;\phi),~~\forall 1 \leq h \leq L.\\
	\end{aligned}
\end{equation}
\begin{equation}
	\label{eq:ims-training-deterministic}
	\phi^\star = \argmin_\phi \mathbb{E}\left[d(\mathbf{\tilde{X}}_{t+1}, m(\mathcal{F}_t;\phi))\right].
\end{equation}
For the probabilistic forecasting model, we use the chain-rule to get the probability of the length-$L$ prediction, which is given in~\eqref{eq:ims-testing-probabilistic}, and estimate the parameters by \emph{Maximum Likelihood Estimation} (MLE).
\begin{equation}
	\label{eq:ims-testing-probabilistic}
	\begin{aligned}
		&p(\mathbf{X}_{t+1:t+L}\mid \mathcal{F}_t;\phi) = p(\mathbf{X}_{t+1} \mid \mathcal{F}_t;\phi) \\
    &\qquad\prod_{i=2}^L p(\mathbf{X}_{t+i} \mid \mathcal{F}_t, \mathbf{X}_{t+1:t+i-1};\phi).\\
	\end{aligned}
\end{equation}
When the one-step-ahead forecaster is probabilistic and nonlinear, the MLE of~\eqref{eq:ims-testing-probabilistic} is generally intractable. In this case, we can draw samples from $p(\mathbf{X}_{t+1:t+L}\mid \mathcal{F}_t;\phi)$ by techniques like \emph{Sequential Monte-Carlo} (SMC)~\cite{del2006sequential} and predict based on these samples.

There are two advantages of the IMS approach: 1) The one-step-ahead forecaster is easy to train because it only needs to consider the one-step-ahead forecasting error and 2) we can generate predictions of an arbitrary length by recursively applying the basic forecaster. However, there is a discrepancy between training and testing in IMS. In the training process, we use the ground-truths of the previous $L-1$ steps to generate the $L$th-step prediction. While in the testing process, we feed back the  model predictions instead of the ground-truths to the forecaster. This makes the model prone to accumulative errors in the forecasting process~\cite{bengio2015scheduled}. For the deterministic forecaster example above, the optimal muti-step-ahead forecaster, which can be obtained by minimizing $\mathbb{E}\left[\sum_{h=1}^L d(\mathbf{\tilde{X}}_{t+h}, m^{(h)}(\mathcal{F}_t;\phi))\right]$, is not the same as recursively applying the optimal one-step-ahead forecaster defined in~\eqref{eq:ims-training-deterministic} when $m$ is nonlinear. This is because the forecasting error at earlier timestamps will propagate to later timestamps~\cite{lin1994forecasting}.

\subsection{Direct Multi-step Estimation}
The main motivation behind DMS is to avoid the error drifting problem in IMS by directly minimizing the long-term prediction error. Instead of training a single model, DMS trains a different model $m_h$ for each forecasting horizon $h$. There can thus be $L$ models in the DMS approach. When the forecasters are deterministic,
the forecasting process and training objective are shown in~\eqref{eq:dms-testing} and~\eqref{eq:dms-training} respectively where $d(\cdot, \cdot)$ is the distance measure.
\begin{equation}
	\label{eq:dms-testing}
	\begin{aligned}
		\mathbf{\hat{X}}_{t+h} &= m_h(\mathcal{F}_t;\phi_L), \forall 1 \leq h \leq L.\\
	\end{aligned}
\end{equation}
\begin{equation}
	\label{eq:dms-training}
	\begin{aligned}
		\phi_1^\star, ..., \phi_L^\star &= \argmin_{\phi_1,...,\phi_L} \mathbb{E}\left[d(\mathbf{\tilde{X}}_{t+1:t+L}, \mathbf{\hat{X}}_{t+1:t+L})\right].
	\end{aligned}
\end{equation}
\begin{equation}
  \label{eq:dms-training-recursive}
  \begin{aligned}
    \phi^\star = \argmin_{\phi} \mathbb{E}\left[\sum_{h=1}^L d(\mathbf{\tilde{X}}_{t+h}, m^{(h)}(\mathcal{F}_t;\phi))\right].
  \end{aligned}
\end{equation}
To disentangle the model size from the number of forecasting steps $L$, we can also construct $m_h$ by recursively applying the one-step-ahead forecaster $m_1$. In this case, the model parameters $\{\phi_1, ..., \phi_L\}$ are shared and the optimization problem turns into~\eqref{eq:dms-training-recursive}. We need to emphasize here that~\eqref{eq:dms-training-recursive}, which optimizes the multi-step-ahead forecasting error, is different from~\eqref{eq:ims-training-deterministic}, which only minimizes the one-step-ahead forecasting error. This construction method is widely adopted in deep learning based methods~\cite{nitish2015unsupervised,oh2015action}.

\subsection{Comparison of IMS and DMS}
Chevillon~\cite{chevillon2007direct} compared the IMS strategy and DMS strategy. According to the paper, DMS can lead to more accurate forecasts when 1) the model is misspecified, 2) the sequences are non-stationary, or 3) the training set is too small. However, DMS is more computationally expensive than IMS. For DMS, if the $\phi_h$s are not shared, we need to store and train $L$ models. If the $\phi_h$s are shared, we need to recursively apply $m_1$ for $O(L)$ steps~\cite{chevillon2007direct, bengio2015scheduled, lamb2016professor}. Both cases require larger memory storage or longer running time than solving the IMS objective. On the other hand, the training process of IMS is easy to parallelize since each forecasting horizon can be trained in isolation from the others~\cite{Goodfellow-et-al-2016-Book}. Moreover, directly optimizing a $L$-step-ahead forecasting loss may fail when the forecasting model is highly nonlinear, or the parameters are not well-initialized~\cite{bengio2015scheduled}.

Due to these trade-offs, models for STSF problems choose the strategies that best match the problems' characteristics. Generally speaking, DMS is preferred for short-term prediction while IMS is preferred for long-term forecast. For example, the IMS strategy is widely adopted in solving TF-MPC problems where the forecasting horizons are long, which are generally above 12~\cite{fragkiadaki2015recurrent, alahi2016social} and can be as long as 100~\cite{jain2016structural}. The DMS strategy is used in STSF-RG problems where the dimensionality of data is usually very large and only short-term predictions are required~\cite{xingjian2015convolutional, mathieu2016deep}. For STSF-IG problems, some works adopt DMS when the overall dimensionality of the forecasting sequence is acceptable for the direct approach~\cite{wytock2013sparse, zheng2015forecasting}.

Overall, IMS is easier to train but less accurate for multi-step forecasting while DMS is more difficult to train but more accurate. Since IMS and DMS are somewhat complementary, later studies have tried to bridge the gap between these two approaches. We will introduce two representative strategies in the following two sections.
\subsection{Boosting Strategy}
\label{ch:boosting}
The boosting strategy proposed in~\cite{taieb2014boosting} combines the IMS and DMS for univariate time series forecasting. The strategy boosts the linear forecaster trained by IMS with several small and nonlinear adjustments trained by DMS. The model assumption is shown in~\eqref{eq:boost-assumption}.
\begin{equation}
	\label{eq:boost-assumption}
	\tilde{x}_{t+h} = m^{(h)}(\mathcal{F}_t;\phi) + \sum_{i=1}^{I_h} \nu l_i(\tilde{x}_{t-j}, \tilde{x}_{t-k};\psi_i) + e_{t,h}.
\end{equation}
Here, $m(\mathcal{F}_t;\phi)$ is the basic one-step-ahead forecasting model,  $l_i(\tilde{x}_{t-j}, \tilde{x}_{t-k};\psi_i)$ is the chosen weak learner at step $i$, $I_h$ is the boosting iteration number, $\nu$ is the shrinkage factor and $e_{t,h}$ is the error term. The author set the base linear learner to be the auto-regressive model and set the weak nonlinear learners to be the penalized regression splines. Also, the weak learners were designed to only account for the interaction between two elements in the observation history. The training method is similar to the gradient boosting algorithm~\cite{friedman2001elements}. $m(\mathcal{F}_t;\phi)$ is first trained by minimizing the IMS objective~\eqref{eq:ims-training-deterministic}. After that, the weak-learners are fit iteratively to the gradient residuals.


The author also performed a theoretical analysis of the bias and variance of different models trained with DMS, IMS and the boosting strategy for 2-step-ahead prediction. The result shows that the estimation bias of DMS can be arbitrarily small while that of the IMS will be amplified for larger forecasting horizon. The model estimated by the boosting strategy also has a bias, but it is much smaller than the IMS approach due to the compensation of the weak learners trained by DMS. Also, the variance of the DMS approach depends on the variability of the input history $\mathcal{F}_t$ and the model size which will be very large for complex and nonlinear models. The variance of the boosting strategy is smaller because the basic model is linear and the weak-learners only consider the interaction between two elements.

Although the author only proposed the boosting strategy for the univariate case, the method, which lies in the mid-ground between IMS and DMS, should apply to the STSF problems with multiple measurements, which leads to potential future works as discussed in Section~\ref{ch:discussion}.
\subsection{Scheduled Sampling}
\label{ch:scheduled-sampling}
The idea of \emph{Scheduled Sampling} (SS)~\cite{bengio2015scheduled} is first to train the model with IMS and then gradually replaces the ground-truths in the objective function with samples generated by the model. When all ground-truth samples are replaced with model-generated samples, the training objective becomes the DMS objective. The generation process of SS is described in~\eqref{eq:ss-training}:
\begin{equation}
  \label{eq:ss-training}
	\begin{aligned}
		&\forall 1 \leq h \leq L,\\
		&\mathbf{\hat{X}}_{t+h} \sim p(\mathbf{X} \mid \mathcal{F}_t, \mathbf{\bar{X}}_{t+1:t+h-1};\phi),\\
		&\mathbf{\bar{X}}_{t+h} = (1 - \tau_{t+h}) \mathbf{\hat{X}}_{t+h} + \tau_{t+h} \mathbf{\tilde{X}}_{t+h},\\
		&\tau_{t+h} \sim B(1, \epsilon_k).
	\end{aligned}
\end{equation}
Here, $\mathbf{\hat{X}}_{t+h}$ and $\mathbf{\tilde{X}}_{t+h}$ are the generated sample and the ground-truth at forecasting horizon $h$, correspondingly. $p(\mathbf{X} \mid \mathcal{F}_t, \mathbf{\bar{X}}_{t+1:t+h-1};\phi)$ is the basic one-step-ahead forecasting model. $\tau_{t+h}$ is a random variable following the binomial distribution, which controls whether to use the ground-truth or the generated sample. $\epsilon_k$ is the probability of choosing the ground-truth at the $k$th iteration. In the training phase, SS minimizes the  distance between $\mathbf{\hat{X}}_{t+1:t+L}$ and $\mathbf{\tilde{X}}_{t+1:t+L}$, which is shown in~\eqref{eq:ss-objective}. In the testing phase, $\tau$ is fixed to 0, meaning that the model-generated samples are always used.
\begin{equation}
\label{eq:ss-objective}
\begin{aligned}
&\mathbb{E}_{p(\mathbf{\hat{X}}_{t+1:t+L}\mid \mathcal{F}_t;\phi)}\big[d(\mathbf{\hat{X}}_{t+1:t+L}, \mathbf{\tilde{X}}_{t+1:t+L}) \big] \\
&= \mathbb{E}_{p(\tau_{t+1:t+L-1})} \big[d(\mathbf{\hat{X}}_{t+1:t+L}, \mathbf{\tilde{X}}_{t+1:t+L})\\
&\qquad\qquad\prod_{h=1}^L p(\mathbf{\hat{X}}_{t+1} \mid \mathcal{F}_t, \mathbf{\bar{X}}_{t+1:t+h-1};\phi)\big].
\end{aligned}
\end{equation}

If $\epsilon_k$ equals to $1$, the ground-truth will always be chosen and the objective function will be the same as in the IMS strategy. If $\epsilon_k$ is $0$, the generated sample will always be chosen and the optimization objective will be the same as in the DMS strategy. Based on this observation, the author proposed to gradually decay $\epsilon_k$ during training to make the optimization objective shift smoothly from IMS to DMS, which is a type of \emph{curriculum learning}~\cite{bengio2009curriculum}. The paper also provided some ways to change the sampling ratio.

One issue of SS is that the expectation over $\tau_{t+h}$s in~\eqref{eq:ss-objective} makes the loss function non-differentiable. The original paper\cite{bengio2015scheduled} obviates the problem by treating $\hat{X}_{t+h}$s as constants, which does not optimize the true objective~\cite{huszar2015not,lamb2016professor}.

The SS strategy is widely adopted in DL models for STSF. In~\cite{finn2016unsupervised}, the author applied SS to video prediction and showed that the model trained via SS outperformed the model trained via IMS. In~\cite{oh2015action}, the author proposed a training strategy similar to SS except that the training curriculum is adapted from IMS to DMS by controlling the number of the prediction steps instead of controlling the frequency of sampling the ground-truth. In~\cite{li2018diffusion,zhang2018gaan}, the author used SS to train models for traffic speed forecasting and showed it outperformed IMS.

\section{Classical Methods for STSF}
\label{ch:method-classical}
In this section, we will review the classical methods for STSF, including feature-based methods, state space models, and Gaussian process models. These methods are generally based on the shallow models or spatiotemporal kernels. Since a spatiotemporal sequence can be viewed as a multivariate sequence if we flatten the $\mathbf{X}_t$s into vectors or treat the observations at every location independently, methods designed for the general multivariate time-series forecasting are naturally applicable to STSF. However, directly doing so ignores the spatiotemporal structure within the data. Therefore, most algorithms introduced in this section have extended this basic approach by utilizing the domain knowledge of the specific task or considering the interactions among different measurements and locations.
\subsection{Feature-based Methods}
\label{ch:feature-based}
Feature-based methods solve the STSF by training a regression model based on some human-engineered spatiotemporal features. To reduce the problem size, most feature-based methods treat the measurements at the $K$ locations to be independent given the extracted features. The high-level IMS and DMS models of the feature-based methods are given in~\eqref{eq:feature-based-ims} and~\eqref{eq:feature-based-dms} respectively.
\begin{align}
	\mathbf{\hat{X}}_{t+1, k} &= f_{\text{IMS}}(\Psi_k(\mathcal{F}_t);\phi) \label{eq:feature-based-ims},\\
	\mathbf{\hat{X}}_{t+1:t+L, k} &= f_{\text{DMS}}(\Psi_k(\mathcal{F}_t;\phi) \label{eq:feature-based-dms}.
\end{align}
Here, $\Psi_k(\mathcal{F}_t)$ represents the features at location $k$. Once the feature extraction method $\Psi_k$ is defined, any regression model $f$ can be applied. The feature-based methods are usually adopted in solving STSF-IG problems. Because the observations are sparsely and irregularly distributed in STSF-IG, it is sometimes more straightforward to extract human-engineered features than to develop a purely machine learning based model. Despite their straightforwardness, the feature-based methods have a major drawback. The model's performance relies heavily on the quality of the human-engineered features. We will briefly introduce two representative feature-based methods for two STSF-IG problems to illustrate the approach.

Ohashi and Torgo~\cite{ohashi2012wind} proposed a feature-based method for wind-speed forecasting. The author designed a set of features called spatiotemporal indicators and applied several regression algorithms including regression tree, support vector regression, and random forest. There are two kinds of spatiotemporal indicators defined in the paper: the average or weighted average of the observed historical values within a space-time neighborhood and the ratio of two average values calculated with different distance thresholds.
The regression model takes the spatiotemporal indicators as the input to generate the one-step-ahead prediction. Experiment results show that the spatiotemporal indicators are helpful for enhancing the forecasting accuracy.

Zheng et al.~\cite{zheng2015forecasting} proposed a feature-based method for air quality forecasting. The author adopted the DMS approach and combined multiple predictors, including the temporal predictor, the spatial predictor, and the inflection predictor, to generate the features at location $k$ and timestamp $t$. The temporal predictor focuses on the temporal side of the data and uses the measurements within a temporal sliding window to generate the features. The spatial predictor focuses on the spatial side of the data and uses the previous measurements within a nearby region to generate the features. Also, to deal with the sparsity and irregularity of the coordinates, the author partitioned the neighborhood region into several groups and aggregated the inner measurements within a group. The inflection predictor extracts the sudden changes within the observations. For the regression model, the temporal predictor uses the linear regression model, the spatial predictor uses a 2-hidden-layer neural network and the inflection predictor used the regression tree model. A regression tree model further combines these predictors.

\subsection{State Space Models}
\label{ch:gsf:statistical-model}
The \emph{State Space Model} (SSM) adopts a generative viewpoint of the sequence. For the observed sequence  $\mathbf{X}_{1:T}$, SSM assumes that each element $\mathbf{X}_t$ is generated by a hidden state $\mathbf{H}_{t}$ and these hidden states, which are allowed to be discrete in our definition, form a Markov process. The general form of the state space model~\cite{murphy2012machine} is given in~\eqref{eq:ssm}. Here, $g$ is the transition model, $f$ is the observation model, $\mathbf{H}_t$ is the hidden state, $\bm{\epsilon}_t$ is the noise of the transition model, and $\bm{\sigma}_t$ is the noise of the observation model. The SSM has a long history in forecasting~\cite{de200625} and many classical time-series models like \emph{Markov Model} (MM), \emph{Hidden Markov Model} (HMM), \emph{Vector Autoregression} (VAR), and \emph{Autoregressive Integrated Moving Average Model} (ARIMA) can be written as SSMs. 
\begin{align}
	\label{eq:ssm}
	\mathbf{H}_t &= g(\mathbf{H}_{t-1}, \bm{\epsilon}_t),\\\nonumber
	\mathbf{X}_t &= f(\mathbf{H}_t, \bm{\sigma}_t).
\end{align}

Given the underlying hidden state $\mathbf{H}_t$, the observation $\mathbf{X}_t$ is independent concerning the historical information $\mathcal{F}_{t}$. Thus, the posterior distribution of the sequence for the next $L$ steps can be written as follows:
\begin{equation}
	\label{eq:ssm:forecasting-posterior}
	\begin{aligned}
	&p(\mathbf{X}_{t+1:t+L}|\mathcal{F}_{t}) = \int p(\mathbf{H}_t \mid \mathcal{F}_t)\\
	&\quad\prod_{i=t+1}^{t+L} p(\mathbf{X}_{i}\mid \mathbf{H}_{i}) p(\mathbf{H}_{i} \mid \mathbf{H}_{i-1}) d\mathbf{H}_{t:t+L}
	\end{aligned}
\end{equation}

The advantage of SSM for solving STSF is that it naturally models the uncertainty of the dynamical system due to its Bayesian formulation. The posterior distribution~\eqref{eq:ssm:forecasting-posterior} encodes the probability of different future outcomes. We can not only get the most likely future but also know the confidence of our prediction by Bayesian inference. Common types of SSMs used in STSF literatures have either discrete hidden states or linear transition models and have limited representational power~\cite{fraccaro2016sequential}. In this section, we will focus on these common types of SSMs and leave the review of more advanced SSMs, which are usually coupled with deep architectures, in Section~\ref{ch:deep-temporal-generative-model}. We will first briefly review the basics of three common classes of SSMs and then introduce their representative spatiotemporal extensions, including the group-based method, \emph{Space-time Autoregressive Moving Average} (STARIMA), and the low-rank tensor auto-regression.


\subsubsection{Common Types of SSMs}
In STSF literature, the following three classes of SSMs are most common.

\textbf{First-order Markov Model:}\quad
When we set $\mathbf{H}_t$ to be $\mathbf{X}_{t-1}$ in~\eqref{eq:ssm} and constrain it to be discrete, the SSM reduces to the first-order MM. Assuming $\mathbf{X}_t$s have $C$ choices, the first-order MM is determined by the transition matrix $\mathbf{A} \in \mathbb{R}^{C \times C}$ where $\mathbf{A}_{i,j} = p(\mathbf{X}_{t+1}=i \mid \mathbf{X}_{t}=j)$ and the prior distribution $\pi \in \mathbb{R}^C$ where $\pi_i = p(\mathbf{X} = i)$. The maximum posterior estimator can be obtained via dynamic programming and the optimal model parameters can be solved in closed form~\cite{murphy2012machine}.

\textbf{Hidden Markov Model:}\quad
When the states are discrete, the SSM turns into the HMM. Assuming the $\mathbf{H}_t$s have $C$ choices, the HMM is determined by the transition matrix $\mathbf{A}\in \mathbb{R}^{C\times C}$, the state prior $\pi$ and the class conditional density $p(\mathbf{X}_t \mid \mathbf{H}_t = i)$. For HMM, the maximum posterior estimator and the optimal model parameters can be calculated via the forward algorithm and the \emph{Expectation Maximization} (EM) algorithm, respectively~\cite{murphy2012machine}.
It is worth emphasizing that the posterior of HMM is tractable even for non-Gaussian observation models. This is because the states of HMM are discrete and it is possible to enumerate all the possibilities.

\textbf{Linear-Gaussian SSM:}\quad
When the states are continuous and both $g$ and $f$ in~\eqref{eq:ssm} are linear-Gaussian, the SSM becomes the \emph{Linear-Gaussian State Space Model} (LG-SSM)~\cite{murphy2012machine}:
\begin{equation}
	\label{eq:lg-ssm}
	\begin{aligned}
		\mathbf{H}_t &= \mathbf{A}_t \mathbf{H}_{t-1} + \bm{\epsilon}_t,\\
		\mathbf{X}_t &= \mathbf{C}_t \mathbf{H}_t + \bm{\sigma}_t,\\
		\bm{\epsilon}_t &\sim \mathcal{N}(0, \mathbf{Q}_t),\\
		\bm{\sigma}_t &\sim \mathcal{N}(0, \mathbf{R}_t),\\
	\end{aligned}
\end{equation}
where $\mathbf{Q}_t$ and $\mathbf{R}_t$ are covariance matrices of the Gaussian noises.
A large number of classical time-series models like VAR and ARIMA can be written in the form of LG-SSM~\cite{murphy2012machine, brodersen2015inferring}. As both the transition model and the observation model are linear-Gaussian, the posterior is also a linear-Gaussian and we can use the \emph{Kalman Filtering} (KF) algorithm~\cite{kalman1960new} to calculate the mean and variance. Similar to HMM, the model parameters can also be learned by EM.

\subsubsection{Group-based Methods}
Group-based methods divide the $N$ locations into non-intersected groups and train an independent SSM for each group. To be more specific, for the location set $S = \{1, 2,..., N\}$, the group-based methods assume that it can be decomposed as $S = \cup_{i=1}^K G_i$ where $G_i \cap G_j = \varnothing,\forall i \neq j$. The measurement sequences that belong to the same group, i.e., $D_i = \{\{\mathbf{X}_{1, j}, ..., \mathbf{X}_{t, j}\}, j \in G_i\}$, are used to fit a single SSM. If $K$ is equal to $1$, it reduces to the simplest case where the measurement sequences at all locations share the same model.

This technique has mainly been applied to TF-MPC problems. Asahara et al.~\cite{asahara2011pedestrian} proposed a group-based extension of MM called \emph{Mixed-Markov-chain Model} (MMM) for pedestrian-movement prediction. MMM assumes that the people belonging to a specific group satisfy the same first-order MM. To use the first-order MM, the author manually discretized the coordinates of each person. The group-assignments along with the model parameters are learned by EM. Experiments show that MMM outperforms MM and HMM in this task. Mathew et al.~\cite{mathew2012predicting} extended the work by using HMM as the base model. Also, in~\cite{mathew2012predicting}, the groups are determined by off-line clustering algorithms instead of EM.

Group-based methods capture the intra-group relationships by model sharing. However, it cannot capture the inter-group correlations. Also, only a single group assignment is considered while there may be multiple grouping relationships in the spatiotemporal data. Moreover, the latent state transitions of SSMs have some internal grouping effects. It is often not necessary to explicitly divide the locations into different groups. Due to these limitations, group-based methods are less mentioned in later research works.

\subsubsection{STARIMA}
STARIMA~\cite{cliff1975model, cliff1975space, pfeifer1980starima} is a classical spatiotemporal extension of the univariate ARIMA model~\cite{box2015time}. STARIMA emphasizes the impact of a location's spatial neighborhoods on its future values. The fundamental assumption is that the future measurements at a specific location $i$ is not only related to the past measurements observed at this location but also related to the past measurements observed at its local neighborhoods. STARIMA is designed for univariate STSF-RG and STSF-IG problems where there is a single measurement at each location. If we denote the measurements at timestamp $t$ as $\mathbf{x}_t \in \mathbb{R}^N$, the model of STARIMA is shown in the following:
\vspace{1em}
\begin{equation}
	\label{eq:starima}
	\begin{aligned}
		\Delta^d \mathbf{x}_t = &\sum_{k=1}^p \sum_{l=0}^{\lambda_k} \phi_{kl} \mathbf{W}^{(l)} \Delta^d \mathbf{x}_{t-k} + \bm{\epsilon}_t\\
    &- \sum_{k=1}^p \sum_{l=0}^{m_k} \theta_{kl} \mathbf{W}^{(l)} \bm{\epsilon}_{t-k}.\\
	\end{aligned}
\end{equation}
Here, $\Delta$ is the temporal difference operator where $\Delta \mathbf{x}_t = \mathbf{x}_t - \mathbf{x}_{t-1}$ and $\Delta ^k \mathbf{x}_t = \Delta \{\Delta^{k-1} \mathbf{x}_t\}$. $\phi_{kl}$ and $\theta_{kl}$ are respectively the autoregressive parameter and the moving average parameter at temporal lag $k$ and spatial lag $l$. $\lambda_k, m_k$ are maximum spatial orders. $\mathbf{W}^{(l)}$ is a predefined $N\times N$ matrix of spatial order $l$. $\bm{\epsilon}_t$ is the normally distributed error vector at time $t$ that satisfies the following relationships:
\vspace{2em}
\begin{equation}
	\begin{aligned}
		\mathbb{E}[\bm{\epsilon}_t] & = \bm{0},\\
		\mathbb{E}\left[\bm{\epsilon}_t \bm{\epsilon}_{t+s}^T\right] & = \begin{cases}
			\sigma^2 \mathbf{I}, &s = 0\\
			\bm{0}, &\text{otherwise}
		\end{cases},\\
		\mathbb{E}\left[\mathbf{x}_t \bm{\epsilon}_{t+s}^T\right] &= 0, \forall s>0.
	\end{aligned}
\end{equation}
\vskip 2em

\begin{figure}[tb!]
	\centering 
	\includegraphics[width=0.47\textwidth]{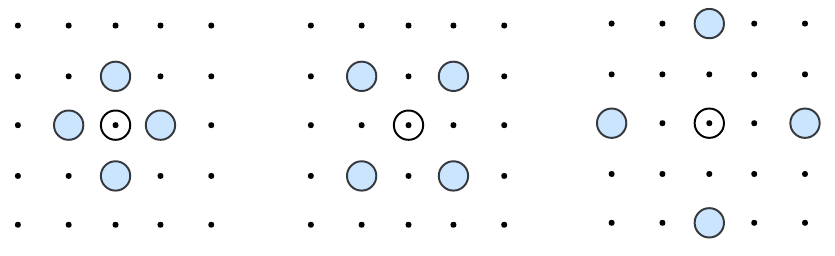}
	\caption{Illustration of the spatial order relationships in STARIMA. From left to right are spatial order equals to 1, 2 and 3. The blue circles are neighbors of the white circle in the middle.}
	\label{fig:starima-spatial-order} 
\end{figure}

The spatial order matrix $\mathbf{W}^{(l)}$ is the key in STARIMA for modeling the spatiotemporal correlations. $\mathbf{W}^{(l)}_{i, j}$ is only non-zero if the site $j$ is the $l$th order neighborhood of site $i$. $\mathbf{W}^{(0)}$ is defined to be the identity matrix. The spatial order relationship in a two-dimensional coordinate system is illustrated in Figure~\ref{fig:starima-spatial-order}. However, the $\mathbf{W}^{(l)}$s need to be predefined by the model builder and cannot vary over-time. Also, the STARIMA still belongs to the class of LG-SSM. It is thus not suitable for modeling nonlinear spatiotemporal systems. Another problem of STARIMA is that it is designed for problems where there is only one measurement at each location and has not considered the case of multiple correlated measurements.

\subsubsection{Low-rank Tensor Auto-regression}
Bahadori et al.~\cite{bahadori2014fast} proposed to train an auto-regression model with low-rank constraint to better capture the latent structure of the spatiotemporal data. This method is designed for STSF-IG and STSF-RG problems where the coordinates are fixed. Assuming the observed measurements $\mathbf{X}_{t} \in \mathbb{R}^{N \times D}$ satisfy the VAR($J$) model, the optimization problem of the low-rank tensor auto-regression is given in \eqref{eq:low-rank-tensor}:
\vspace{1em}
\begin{equation}
	\begin{aligned}
  \label{eq:low-rank-tensor}
		\min_{\mathcal{W}}&\sum_{t=J+1}^T \norm{\hat{\mathbf{X}}_t - \mathbf{X}_t}_F^2 + \mu \sum_{t=1}^T \text{tr}(\hat{\mathbf{X}}_t^T \mathbf{L} \hat{\mathbf{X}}_t),\\
		s.t.&\quad r(\mathcal{W}) \leq \rho,\\
		\quad&\quad\hat{\mathbf{X}}_{t + 1, k} = \mathcal{W}_{k}\mathbf{X}_{t-J+1:t},~\forall t, k.
	\end{aligned}
\end{equation}
Here, $\mathcal{W} \in \mathbb{R}^{N \times D \times JD}$ is the three-dimensional weight tensor. $\mathbf{L} \in \mathbb{R}^{D \times D}$ is the spatial Laplacian matrix calculated from the fixed coordinate matrix $\mathbf{C} \in \mathbb{R}^{N \times E}$. $r(\mathcal{W})$ means the rank of the weight tensor and $\rho$ is the maximum rank. The regularization term in the objective function constrains the predicted results to be spatially smooth. $\mu$ is the strength of this spatial similarity constraint.

Constraining the rank of the weight tensor imposes an implicit spatiotemporal structure in the data. When the weight tensor is low-rank, the observed sequences can be described with a few latent factors.

The optimization problem~\eqref{eq:low-rank-tensor} is non-convex and NP-hard in general~\cite{recht2010guaranteed} due to the low-rank constraint. One standard approach is to relax the rank constraint to a nuclear norm constraint and optimize the relaxed problem. However, this approach is not computationally feasible for solving~\eqref{eq:low-rank-tensor} because of the high dimensionality of the spatiotemporal data. Thus, researchers~\cite{bahadori2014fast, yu2015accelerated, yu2016learning} proposed optimization algorithms to accelerate the learning process.

The primary limitation of the low-rank tensor auto-regression is that the regression model is still linear even if we have perfectly learned the weights. We cannot directly apply it to solve forecasting problems that have strong nonlinearities.

\subsection{Gaussian Process}
\emph{Gaussian Process} (GP) is a non-parametric Bayesian model. GP defines a prior $p(f)$ in the function space and tries to infer the posterior $p(f\mid D)$ given the observed data~\cite{murphy2012machine}. GP assumes the joint distribution of the function values of an arbitrary finite set $\{x_1, x_2,..., x_n\}$, i.e., $p(f(x_1), ..., f(x_n))$, is a Gaussian distribution. A typical form of the GP prior is shown in~\eqref{eq:gp-prior}:
\begin{equation}
	\label{eq:gp-prior}
	f \sim \mathcal{GP}(m(x), \kappa(x, x^\prime)),
\end{equation}
where $m(x)$ is the mean function and $\kappa(x, x^\prime)$ is the kernel function.

In geostatistics, GP regression is also known as kriging~\cite{journel1978mining,rasmussen2006gaussian}, which is used to interpolate the missing values of the data. When applied to STSF, GP is usually used in STSF-IG and STSF-RG problems where the coordinates are fixed. The general formula of the GP based models is shown in~\eqref{eq:gp:type1}:
\begin{equation}
	\label{eq:gp:type1}
	\begin{aligned}
		f &\sim \mathcal{GP}(m(X), \kappa_\theta),\\
		Y &\sim p(Y \mid f(X)).
	\end{aligned}
\end{equation}
Here, $X$ contains the space-time positions of the samples, which is the concatenation of the spatial coordinates and the temporal index. For example, $X_k = (t_k, i_k, j_k)$ where $t_k$ is the timestamp and $(i_k, j_k)$ is the spatial coordinate. $\kappa_\theta$ is the kernel function parameterized by~$\theta$. $p(Y \mid f(X))$ is the observation model. Like SSM, GP is also a generative model. The difference is that GP directly generates the function describing the relationship between the observation and the future while SSM generates the future measurements one-by-one through the Markov assumption. We need to emphasize here that GP can also be applied to solve TF-MPC problems. A well-known example is the \emph{Interacting Gaussian Processes} (IGP)~\cite{trautman2010unfreezing}, in which a GP models the movement of each person, and a potential function defined over the positions of different people models their interactions. Nevertheless, the method for calculating the posterior of IGP is similar to that of~\eqref{eq:gp:type1} and we will stick to this model formulation.

To predict the measurements given a GP model, we are required to calculate the predictive posterior $p(f^{\star} \mid X^{\star}, X^{\mathcal{D}}, Y^{\mathcal{D}})$ where $\{X^{\mathcal{D}}, Y^{\mathcal{D}}\}$ are the observed data points and $X^{\star}$ is the candidate space-time coordinates to infer the measurements. If the observation model is a Gaussian distribution, i.e., $p(Y \mid f(X)) = \mathcal{N}(f(X), \sigma^2)$, the posterior distribution is also a Gaussian distribution as shown in~\eqref{eq:gp_normal}.
\begin{equation}
\label{eq:gp_normal}
  \begin{aligned}
    &p(f^{\star} \mid X^{\star}, X^{\mathcal{D}}, Y^{\mathcal{D}}) = \mathcal{N}(\bm{\mu}^\star, \bm{\Sigma}^\star),\\
    &\mathbf{K} = \kappa(X^{\mathcal{D}}, X^{\mathcal{D}}) + \sigma^2 \mathbf{I}\\
    &\bm{\mu}^\star =\kappa(X^{\star}, X^{\mathcal{D}})\mathbf{K}^{-1} Y^{\mathcal{D}},\\
    &\bm{\Sigma}^\star = \kappa(X^\star, X^\star) - \kappa(X^\star, X^{\mathcal{D}}) \mathbf{K}^{-1} \kappa(X^\star, X^{\mathcal{D}}),
  \end{aligned}
\end{equation}
If the observation model is non-Gaussian, which is common for STSF problems~\cite{flaxman2015machine}, Laplacian approximation can be utilized to approximate the posterior~\cite{murphy2012machine}. The inference process of GP is computationally expensive due to the involvement of $\mathbf{K}^{-1}$ and the naive approach requires $O(\abs{D}^3)$ computations and $O(\abs{D}^2)$ storage. There are many acceleration techniques to speed up the inference~\cite{flaxman2015machine}. We will not investigate them in detail in this survey and readers can refer to~\cite{rasmussen2006gaussian,flaxman2015machine}.

The key ingredient of GP is the kernel function. For STSF problems, the kernel function should take both the spatial and temporal correlations into account. The common strategy for defining complex kernels is to combine multiple basic kernels by operations like summation and multiplication. Thus, GP models for STSF generally use different kernels to model different aspects of the spatiotemporal sequence and ensemble them together to form the final kernel function. In~\cite{flaxman2015machine}, the author proposed to decompose the spatiotemporal kernel as $\kappa_{st}((s,t), (s^\prime, t^\prime)) = \kappa_s(s, s^\prime) \kappa_t(t, t^\prime)$ where $\kappa_s$ is the spatial kernel and $\kappa_t$ is the temporal kernel~\cite{flaxman2015machine}. If we denote the kernel matrix corresponding to $\kappa_{st}, \kappa_s, \kappa_t$ as $\mathbf{K}_{st}, \mathbf{K}_{s}, \mathbf{K}_{t}$, this decomposition results in $\mathbf{K}_{st} = \mathbf{K}_{s} \otimes \mathbf{K}_{t}$ where $\otimes$ is the Kronecker product. Based on this observation, the author utilized the computational benefits of Kronecker algebra to scale up the inference of the GP models~\cite{flaxman2015fast}. In~\cite{senanayake2016predicting}, the author combined two spatial kernels and one temporal kernel to construct a GP model for seasonal influenza prediction. The author defined the basic spatial kernels as the average of multiple RBF kernels. The temporal kernel was defined to be the sum of the periodic kernel, the Paciorek’s kernel, and the exponential kernel. The final kernel function used in the paper was $\kappa_{\text{final}} = \kappa_{\text{space}} + \kappa_{\text{time}} + \kappa_{\text{time}} \times \kappa_{\text{space2}}$.
\subsection{Remarks}
\label{ch:classical-remark}
In this section, we reviewed three types of classical methods for solving the STSF problems, feature-based methods, SSMs, and GP models. The feature-based methods are simple to implement and can provide useful predictions in some practical situations. However, these methods rely heavily on the human-engineered features and may not be applied without the help of domain experts. The SSMs assume that the observations are generated by Markovian hidden states. We introduced three common types of SSMs: first-order MM, HMM and LG-SSM. Group-based methods extend SSMs by creating groups among the states or observations. STARIMA extends ARIMA by explicitly modeling the impact of nearby locations on the future. Low-rank tensor autoregression extends the basic autoregression model by adding low-rank constraints on the model parameters. The advantage of these common types of SSMs for STSF is that we can calculate the exact posterior distribution and conduct Bayesian inference. However, the overall nonlinearity of these models are limited and may not be suitable for some complex STSF problems like video prediction. GP models the inner characteristics of the spatiotemporal sequences by spatiotemporal kernels. Although being very powerful and flexible, GP has high computational and storage complexity. Without any optimization, the inference of GP requires $O(\abs{D}^3)$ computations and  $O(\abs{D}^2)$ storage, which makes it not suitable for cases when a huge number of training instances are available.

\section{Deep Learning Methods for STSF}
\label{ch:dl}
\emph{Deep Learning} (DL) is an emerging new area in machine learning that studies the problem of how to learn a hierarchically-structured model to directly map the raw inputs to the desired outputs~\cite{Goodfellow-et-al-2016-Book}. Usually, DL model stacks some basic learnable blocks, or layers, to form a deep architecture. The overall network is then trained end-to-end. In recent years, a growing number of researchers have proposed DL models for STSF. To better understand these models, we will start with an introduction of the relevant background of DL and then review the models for STSF.

\subsection{Preliminaries}
\subsubsection{Basic Building Blocks}
\label{ch:basic-building-blocks}
In this section, we will review the basic blocks that have been used to build DL models for STSF. Since there is a vast literature on DL, the introduction of this part serves only as a brief introduction. Readers can refer to~\cite{Goodfellow-et-al-2016-Book} for more details.

\textbf{Restricted Boltzmann Machine:}\quad
\emph{Restricted Boltzmann Machine} (RBM)~\cite{smolensky1986information} is the basic building block of \emph{Deep Generative Models} (DGMs). RBMs are undirected graphical models that contain a layer of observable variables $\mathbf{v}$ and another layer of hidden variables $\mathbf{h}$. These two layers are interconnected and form a bipartite graph. In its simplest form, $\mathbf{h}$ and $\mathbf{v}$ are all binary and contain $n_h$ and $n_v$ nodes. The joint distribution and the conditional distributions are defined in~\eqref{eq:rbm_basic}~\cite{Goodfellow-et-al-2016-Book}:
\begin{equation}
  \label{eq:rbm_basic}
	\begin{aligned}
		p(\mathbf{v}, \mathbf{h}) &= \frac{1}{Z} \exp(-E(\mathbf{v}, \mathbf{h})),\\
		E(\mathbf{v}, \mathbf{h}) &= -\mathbf{b}^T \mathbf{v} - \mathbf{c}^T \mathbf{h} - \mathbf{v}^T \mathbf{W} \mathbf{h},\\
		p(\mathbf{h} \mid \mathbf{v}) &= \prod_{j=1}^{n_h} \sigma\left((2\mathbf{h}-1)\circ(\mathbf{c} + \mathbf{W}^T\mathbf{v})\right)_j,\\
		p(\mathbf{v} \mid \mathbf{h}) &= \prod_{j=1}^{n_v} \sigma\left((2\mathbf{v}-1)\circ(\mathbf{b} + \mathbf{W}\mathbf{h})\right)_j.
	\end{aligned}
\end{equation}
Here, $\mathbf{b}$ is the bias of the visible state, $\mathbf{c}$ is the bias of the hidden state, $\mathbf{W}$ is the weight between visible states and hidden states, $\sigma$ is the sigmoid function defined in Table~\ref{tbl:activations}, and $Z$ is the normalization constant for energy-based models, which is also known as the partition function~\cite{murphy2012machine}. RBM can be learned by Contrastive Divergence (CD)~\cite{hinton2002training}, which approximates the gradient of $\log Z$ by drawing samples from $k$ cycles of MCMC.

The basic form of RBM has been extended in various ways. \emph{Gaussian-Bernoulli RBM} (GRBM)~\cite{welling2004exponential} assumes the visible variables are continuous and replaces the Bernoulli distribution with a Gaussian distribution. \emph{Spike and Slab RBM} (ssRBM) maintains two types of hidden variables to model the covariance.

\textbf{Activations:}\quad
Activations refer to the element-wise nonlinear mappings $\mathbf{h}=f(\mathbf{x})$. Activation functions that are used in the reviewed models are listed in Table~\ref{tbl:activations}.
\begin{table}[tb!]
  \centering
  \caption{Activations that appear in models reviewed in this survey.}
  \label{tbl:activations}
  \begin{tabular}{|l|l|}
    \hline
    \textbf{Name}               & \textbf{Formula}                                 \\ \hline
    $\sigma$ (sigmoid) & $\sigma(x) = \frac{1}{1 + e^{-x}}$      \\ \hline
    tanh               & $\text{tanh}(x) = \frac{e^x - e^{-x}}{e^x + e^{-x}}$ \\ \hline
    ReLU~\cite{glorot2011deep}         & $\text{ReLU}(x) = \max(0, x)$           \\ \hline
    Leaky ReLU~\cite{maas2013rectifier} & $\text{LReLU}(x) = \max(-\alpha x, x)$ \\ \hline
  \end{tabular}
\end{table}

\textbf{Sigmoid Belief Network:}\quad
\emph{Sigmoid Belief Network} (SBN) is another building block of DGM. Unlike RBM which is undirected, SBN is a directed graphical model. SBN contains a hidden binary state $\mathbf{h}$ and a visible binary state $\mathbf{v}$. The basic generation process of SBN is given in the following:
\begin{equation}
	\begin{aligned}
		p(\mathbf{v}_m \mid \mathbf{h}) &= \sigma(\mathbf{w}_m^T \mathbf{h} + \mathbf{c}_m),\\
		p(\mathbf{h}_j) &= \sigma(\mathbf{b}_j).
	\end{aligned}
\end{equation}
Here, $\mathbf{b}_j, \mathbf{c}_m$ are biases and $\mathbf{w}_m$ is the weight. Because the posterior distribution is intractable, SBN is learned by methods like \emph{Neural Variational Inference and Learning} (NVIL)~\cite{MnihGregor2014} and \emph{Wake-Sleep} (WS)~\cite{hinton1995wake} algorithm, which train an inference network along with the generation network.

\textbf{Fully-connected Layer:}\quad
The \emph{Fully-Connected} (FC) layer is the very basic building block of DL models. It refers to the linear-mapping from the input $\mathbf{x}$ to the hidden state $\mathbf{h}$, i.e, $\mathbf{h} = \mathbf{W} \mathbf{x} + \mathbf{b}$, where $\mathbf{W}$ is the weight and $\mathbf{b}$ is the bias.

\textbf{Convolution and Pooling:}\quad
The convolution layer and pooling layer are first proposed to take advantage of the translation invariance property of image data. The convolution layer computes the output by scanning over the input and applying the same set of linear filters. Although the input can have an arbitrary dimensionality~\cite{tran2015learning}, we will just introduce 2D convolution and 2D pooling in this section. For input $\mathcal{X}\in \mathbb{R}^{C_i\times H_i \times W_i}$, the output of the convolution layer $\mathcal{H}\in \mathbb{R}^{C_o\times H_o \times W_o}$, which is also known as feature map, will be calculated as the following:
\begin{equation}
	\label{eq:convolution-layer-original}
	\mathcal{H} = \mathcal{W} \ast \mathcal{X} + \mathbf{b}.
\end{equation}
Here, $\mathcal{W} \in \mathbb{R}^{C_i \times C_o \times K_h \times K_w}$ is the weight, $K_h$ and $K_w$ are the kernel sizes, ``$\ast$'' denotes convolution, and $\mathbf{b}$ is the bias. If we denote the feature vector at the $(i, j)$ position as $\mathcal{H}_{:, i, j}$, the flattened weight as $\mathbf{W} = \text{vec}(\mathcal{W})$, and the flattened version of the local region inputs as $\mathbf{x}_{\mathcal{N}(i, j)} = \text{vec}([\mathcal{X}_{:, s, t} \mid (s, t) \in \mathcal{N}(i, j)])$, \eqref{eq:convolution-layer-original} can be rewritten as~\eqref{eq:convolution-layer-fc}:
\begin{equation}
	\label{eq:convolution-layer-fc}
	\mathcal{H}_{:, i, j} = \mathbf{W} \mathbf{x}_{\mathcal{N}(i, j)} + \mathbf{b}.
\end{equation}
This shows that convolution layer can also be viewed as applying multiple FC layers with shared weight and bias on local regions of the input. Here, the ordered neighborhood set $\mathcal{N}(i, j)$ is determined by hyper-parameters like the kernel size, stride, padding, and dilation~\cite{yu2016multi}. For example, if the kernel size of the convolution filter is $3 \times 3$, we have $\mathcal{N}(i, j) = \{(i + s, j + t) \mid -1 \leq s, t \leq 1\}$. Readers can refer to~\cite{Goodfellow-et-al-2016-Book,yu2016multi} for details about how to calculate the neighborhood set $\mathcal{N}(i, j)$ based on these hyper-parameters. Also, the out-of-boundary regions of the input are usually set to zero, which is known as zero-padding.

Like convolution layer, the pooling layer also scans over the input and applies the same mapping function. Pooling layer generally has no parameter and uses some reduction operations like $\text{max}$, $\text{sum}$, and $\text{average}$ for mapping. The formula of the pooling layer is given in the following:
\begin{equation}
	\label{eq:pooling-layer-fc}
	\mathcal{H}_{k, i, j} = g(\{\mathcal{X}_{k, s, t} \mid (s, t) \in \mathcal{N}(i, j)\}),
\end{equation}
where $g$ is the element-wise reduction operation.

\textbf{Deconvolution and Unpooling:}\quad
The deconvolution~\cite{zeiler2010deconvolutional} and unpooling layer~\cite{zeiler2014visualizing} are first proposed to serve as the ``inverse'' operation of the convolution and the pooling layer. Computing the forward pass of these two layers is equivalent to computing the backward pass of the convolution and pooling layers~\cite{zeiler2010deconvolutional}.

\textbf{Graph Convolution:}\quad
Graph convolution generalizes the standard convolution, which is operated over a regular grid topology, to convolution over graph structures~\cite{bruna2014spectral,duvenaud2015convolutional,kipf2017semi,zhang2018gaan}. There are two interpretations of graph convolution: the spectral interpretation and the spatial interpretation. In the spectral interpretation~\cite{bruna2014spectral,defferrard2016convolutional}, graph convolution is defined by the convolution theorem, i.e., the Fourier transform of convolution is the point-wise product of Fourier transforms:
\begin{equation}
\label{eq:graph_conv}
\begin{aligned}
\mathbf{x} \ast_G \mathbf{y} &= F^{-1}(F(\mathbf{x}) \circ F(\mathbf{y})) = \mathbf{U}(\mathbf{U}^T \mathbf{x} \circ \mathbf{U}^T \mathbf{y}).
\end{aligned}
\end{equation}
Here, $F(\mathbf{x}) = \mathbf{U}^T \mathbf{x}$ is the graph Fourier transform. The transformation matrix $\mathbf{U}$ is defined based on the eigendecomposition of the graph Laplacian. Defferrard et al.~\cite{defferrard2016convolutional} defined $U$ as the eigenvectors of $\mathbf{I} - \mathbf{D}^{-\frac{1}{2}}\mathbf{A}\mathbf{D}^{-\frac{1}{2}}$, in which $\mathbf{D}$ is the diagonal degree matrix, $\mathbf{A}$ is the adjacency matrix, and $\mathbf{I}$ is the identity matrix.

In the spatial interpretation, graph convolution is defined by a localized parameter-sharing operator that aggregates the features of the neighboring nodes, which can also be termed as graph aggregator~\cite{hamilton2017inductive,zhang2018gaan}:
\begin{equation}
\label{eq:graph_conv_spatial}
\begin{aligned}
\mathbf{h}_i = \gamma_\theta(\mathbf{x}_i, \{\mathbf{z}_{\mathcal{N}_i}\}),
\end{aligned}
\end{equation}
where $\mathbf{h}_i$ is the output feature vector of node $i$, $\gamma_\theta(\cdot)$ is the mapping function parameterized by $\theta$, $\mathbf{x}_i$ is the input feature of node $i$, and $\mathbf{z}_{\mathcal{N}_i}$ contains the features of node $i$'s neighborhoods. The mapping function needs to be permutation invariant and dynamically resizing. One example is the pooling aggregator defined in~\eqref{eq:graph_conv_pool}:
\begin{equation}
\label{eq:graph_conv_pool}
\begin{aligned}
\mathbf{h}_i  &= \phi_o(\mathbf{x}_i \oplus \text{pool}_{j \in \mathcal{N}_i}(\phi_v(\mathbf{z}_j))),
\end{aligned}
\end{equation}
where $\phi_o$ and $\phi_v$ are mapping functions and $\oplus$ means concatenation. Compared with the spectral interpretation, the spatial interpretation does not require the expensive eigendecomposition and is more suitable for large graphs. Moreover, Defferrard et al.~\cite{defferrard2016convolutional} and Kipf \& Welling~\cite{kipf2017semi} proposed accelerated versions of the spectral graph convolution that could be interpreted from the spatial perspective.


\subsubsection{Types of DL Models for STSF}
Based on the characteristic of the model formulation, we can categorize DL models for STSF into two major types: \emph{Deep Temporal Generative Models} (DTGMs) and \emph{Feedforward Neural Network} (FNN) and \emph{Recurrent Neural Network} (RNN) based methods. In the following, we will briefly introduce the basics about DTGM, FNN, and RNN.

\textbf{DTGM:}\quad
DTGMs are DGMs for temporal sequences. DTGM adopts a generative viewpoint of the sequence and tries to define the probability distribution $p(\mathbf{X}_{1:T})$. The aforementioned RBM and SBN are building-blocks of DTGM. 

\textbf{FNN:}\quad
FNNs, also called deep feedforward networks~\cite{Goodfellow-et-al-2016-Book}, refer to deep learning models that construct the mapping $\mathbf{y} = f(\mathbf{x};\theta)$ by stacking various of the aforementioned basic blocks such as FC layer, convolution layer, deconvolution layer, graph convolution layer, and activation layer. Common types of FNNs include \emph{Multilayer Perceptron} (MLP)~\cite{rosenblatt1962perceptions,Goodfellow-et-al-2016-Book}, which stacks multiple FC layers and nonlinear activations, \emph{Convolutional Neural Network} (CNN)~\cite{alex2012imagenet}, which stacks multiple convolution and pooling layers, and graph CNN~\cite{defferrard2016convolutional}, which stacks multiple graph convolution layers. The parameters of FNN are usually estimated by minimizing the loss function plus some regularization terms.
Usually, the optimization problem is solved via stochastic gradient-based methods~\cite{Goodfellow-et-al-2016-Book}, in which the gradient is computed by backpropagation~\cite{nocedal2006numerical}.

\textbf{RNN:}\quad
RNN generalizes the structure of FNN by allowing cyclic connections in the network. These recurrent connections make RNN especially suitable for modeling sequential data. Here, we will only introduce the discrete version of RNN which updates the hidden states using $\mathbf{h}_t = f(\mathbf{h}_{t-1}, \mathbf{x}_t;\theta)$~\cite{Goodfellow-et-al-2016-Book}. These hidden states can be further used to compute the output and define the loss function. One example of an RNN model is shown in~\eqref{eq:vanilla-rnn}. The model uses the $\text{tanh}$ activation and the fully-connected layer for recurrent connections. It uses another full-connected layer to get the output:
\begin{equation}
	\label{eq:vanilla-rnn}
  \begin{aligned}
	\mathbf{h}_t &= \text{tanh}(\mathbf{W}_h \mathbf{h}_{t-1} + \mathbf{W}_x \mathbf{x}_t + b),\\
	\mathbf{o}_t &= \mathbf{W}_o \mathbf{h}_t + \mathbf{c}.
  \end{aligned}
\end{equation}
Similar to FNNs, RNNs are also trained by stochastic gradient based methods. To calculate the gradient, we can first unfold an RNN to a FNN and then perform backpropagation on the unfolded computational graph. This algorithm is called \emph{Backpropagation Through Time} (BPTT)~\cite{Goodfellow-et-al-2016-Book}.

Directly training the RNN shown in~\eqref{eq:vanilla-rnn} with BPTT will cause the vanishing and exploding gradient problems~\cite{pascanu2013difficulty}. One way to solve the vanishing gradient problem is to use the \emph{Long-Short Term Memory} (LSTM)~\cite{hochreiter1997long}, which has a cell unit that is designed to capture the long-term dependencies. The formula of LSTM is given in the following:
\begin{equation}
	\begin{aligned}
		\mathbf{i}_t &= \sigma(\mathbf{W}_{xi} \mathbf{x}_t + \mathbf{W}_{hi} \mathbf{h}_{t-1} + \mathbf{b}_i),\\
		\mathbf{f}_t &= \sigma(\mathbf{W}_{xf} \mathbf{x}_t + \mathbf{W}_{hf} \mathbf{h}_{t-1} + \mathbf{b}_f ),\\
		\mathbf{c}_t &= \mathbf{f}_t \circ \mathbf{c}_{t-1} +\\
    &\qquad \mathbf{i}_t \circ \text{tanh}(\mathbf{W}_{xc} \mathbf{x}_t + \mathbf{W}_{hc} \mathbf{h}_{t-1} + \mathbf{b}_c),\\
		\mathbf{o}_t &= \sigma(\mathbf{W}_{xo} \mathbf{x}_t + \mathbf{W}_{ho} \mathbf{h}_{t-1} + \mathbf{b}_o),\\
		\mathbf{h}_t &= \mathbf{o}_t \circ \text{tanh}(\mathbf{c}_t).
	\end{aligned}
\end{equation}
Here, $\mathbf{c}_t$ is the memory cell and $\mathbf{i}_t, \mathbf{f}_t, \mathbf{o}_t$ are the input gate, the forget gate, and the output gate, respectively. The memory cell is accessed by the next layer when $\mathbf{o}_t$ is turned on. Also, the cell can be updated or cleared when $\mathbf{i}_t$ or $\mathbf{f}_t$ is activated. By using the memory cell and gates to control information flow, the gradient will be trapped in the cell and will not vanish over time, which is also known as the \emph{Constant Error Carousel} (CEC)~\cite{hochreiter1997long}. LSTM belongs to a broader category of RNNs called gated RNNs~\cite{Goodfellow-et-al-2016-Book}. We will not introduce other architectures like \emph{Gated Recurrent Unit} (GRU) in this survey because LSTM is the most representative example. To simplify the notation, we will also denote the transition rule of LSTM as
$\mathbf{h}_t, \mathbf{c}_t = \text{LSTM}(\mathbf{h}_{t-1}, \mathbf{c}_{t-1}, \mathbf{x}_t; \mathbf{W}).$
\subsection{Deep Temporal Generative Models}
\label{ch:deep-temporal-generative-model}
\begin{figure}[tb]
	\centering
	\includegraphics[width=0.5\textwidth]{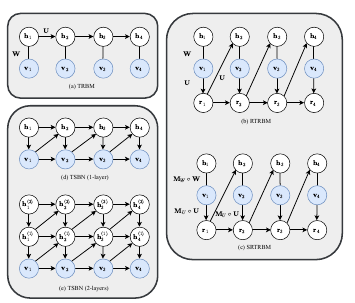}
	\caption{Structure of the surveyed four types of deep temporal generative models: TRBM, RTRBM, SRTRBM and TSBN.}
	\label{fig:dtgm-structures}
\end{figure}

In this section, we will review DTGMs for STSF. Similar to SSM and GP model, DTGM also adopts a generative viewpoint of the sequences. The characteristics of DTGMs is that models belonging to this class can all be stacked to form a deep architecture, which enables them to model complex system dynamics. We will survey four DTGMs in the following. The general structures of these models are shown in Figure~\ref{fig:dtgm-structures}.


\textbf{Temporal Restricted Boltzmann Machine:}\quad
Sutskever \& Hinton~\cite{sutskever2007learning} proposed \emph{Temporal Restricted Boltzmann Machine} (TRBM), which is a temporal extension of RBM. In TRBM, the bias of the RBM in the current timestamp is determined by the previous $m$ RBMs. The conditional probability of TRBM is given in the following:
\begin{equation}
	\begin{aligned}
		p(\mathbf{v}_t, \mathbf{h}_t \mid \mathbf{\hat{v}}_t, \mathbf{\hat{h}}_{t}) &= \frac{1}{Z}\exp\left(-E(\mathbf{v}_t, \mathbf{h}_t \mid \mathbf{\hat{v}}_{t}, \mathbf{\hat{h}}_{t})\right),\\
		E(\mathbf{v}_t, \mathbf{h}_t \mid \mathbf{\hat{v}}_{t}, \mathbf{\hat{h}}_{t}) &=
     -f_{b_v}(\mathbf{\hat{v}}_{t})^T \mathbf{v}_t -f_{b_h}(\mathbf{\hat{v}}_{t}, \mathbf{\hat{h}}_{t})^T \mathbf{h}_t \\
     &~~~~- \mathbf{v}_t^T\mathbf{W}\mathbf{h}_t,\\
		f_{b_v}(\mathbf{\hat{v}}_{t}) &= \mathbf{a} + \sum_{i=1}^m \mathbf{A}_i^T \mathbf{v}_{t-i},\\
		f_{b_h}(\mathbf{\hat{v}}_{t}, \mathbf{\hat{h}}_{t}) =& \mathbf{b} + \sum_{i=1}^m \mathbf{B}_i^T \mathbf{v}_{t-i} + \sum_{i=1}^m \mathbf{C}_i^T \mathbf{h}_{t-i},
	\end{aligned}
\end{equation}
where $\mathbf{\hat{v}}_t = \mathbf{v}_{t-m:t-1}$ and $\mathbf{\hat{h}}_t = \mathbf{h}_{t-m:t-1}$.
The joint distribution is modeled as the multiplication of the conditional probabilities, i.e., $p(\mathbf{v}_{1:T}, \mathbf{h}_{1:T}) = \prod_{i=1}^T p(\mathbf{v}_t, \mathbf{h}_t \mid \mathbf{\hat{v}}_{t}, \mathbf{\hat{h}}_{t})$.

The inference of TRBM is difficult and involves evaluating the ratio of two partition functions~\cite{sutskever2009recurrent}. The original paper adopted a heuristic inference procedure~\cite{sutskever2007learning}. The results on a synthetic bouncing ball dataset show that a two-layer TRBM outperforms a single-layer TRBM in this task.

\textbf{Recurrent Temporal Restricted Boltzmann Machine:}\quad
Later, Sutskever et al.~\cite{sutskever2009recurrent} proposed the \emph{Recurrent Temporal Restricted Boltzmann Machine} (RTRBM) that extends the idea of TRBM by conditioning the parameters of the RBM at timestamp $t$ on the output of an RNN. The model of RTRBM is described in the following:
\begin{equation}
	\begin{aligned}
		p(\mathbf{v}_t, \mathbf{h}_t \mid \mathbf{r}_{t-1}) &= \frac{1}{Z}\exp\left(-E(\mathbf{v}_t, \mathbf{h}_t \mid \mathbf{r}_{t-1})\right),\\
		E(\mathbf{v}_t, \mathbf{h}_t \mid \mathbf{r}_{t-1}) &= -\mathbf{h}_t^T \mathbf{W} \mathbf{v}_t - \mathbf{c}^T \mathbf{v}_t - \mathbf{b}^T \mathbf{h}_t \\
    &~~~~- \mathbf{h}_t^T \mathbf{U} \mathbf{r}_{t-1},\\
		\mathbf{r}_t &= \begin{cases}
			\sigma(\mathbf{W} \mathbf{v}_t + \mathbf{U} \mathbf{r}_{t-1} + \mathbf{b}),& t > 1\\
			\sigma(\mathbf{W} \mathbf{v}_t + \mathbf{b}_{init}),& t = 1
		\end{cases}.
	\end{aligned}
\end{equation}
Here, $\mathbf{r}_{t}$ encodes the visible states $\mathbf{v}_{1:t}$.
The inference is simpler in RTRBM than TRBM because the model is the same as the standard RBM once $\mathbf{r}_{t-1}$ is given. Similar to RBM, RTRBM can also be learned by CD. The author also extended RTRBM to use GRBM for modeling continuous time-series. Qualitative experiments on the synthetic bouncing ball dataset show that RTRBM can generate more realistic sequences than TRBM.

\textbf{Structured Recurrent Temporal Restricted Boltzmann Machine:}\quad
Mittelman et al.~\cite{mittelman2014structured} proposed \emph{Structured Recurrent Temporal Restricted Boltzmann Machine} (SRTRBM) that improves upon RTRBM by learning the connection structure based on the spatiotemporal patterns of the input. SRTRBM uses two block-masking matrices  to mask the weights between the hidden and visible nodes. The formula of SRTRBM is similar to RTRBM except for the usage of two masking matrices $\mathbf{M}_W$ and $\mathbf{M}_U$:
\begin{equation}
	\begin{aligned}
		&E(\mathbf{v}_t, \mathbf{h}_t \mid \mathbf{r}_{t-1}) = -\mathbf{h}_t^T\mathbf{\hat{W}} \mathbf{v}_t - \mathbf{c}^T \mathbf{v}_t\\
     &\qquad\qquad\qquad\quad~~~~- \mathbf{b}^T \mathbf{h}_t - \mathbf{h}_t^T \mathbf{\hat{U}} \mathbf{r}_{t-1},\\
		&\mathbf{r}_t = \begin{cases}
			\sigma(\mathbf{\hat{W}} \mathbf{v}_t + \mathbf{\hat{U}} \mathbf{r}_{t-1} + \mathbf{b}),& t > 1\\
			\sigma(\mathbf{\hat{W}} \mathbf{v}_t + \mathbf{b}_{init}),& t = 1
		\end{cases},\\
    &\mathbf{\hat{W}} = \mathbf{W} \circ \mathbf{M}_W,\\
    &\mathbf{\hat{U}} = \mathbf{U} \circ \mathbf{M}_U.
	\end{aligned}
\end{equation}
Here, $\mathbf{M}_W$ and $\mathbf{M}_U$ are block-masking matrices generated by a graph $G = (V;E)$. To construct the graph, the visible units and the hidden units are divided into $\abs{V}$ nonintersecting subsets. Each node $\epsilon$ in the graph is assigned to a subset of hidden units $C_\epsilon^h$ and a subset of visible units $C_\epsilon^v$. The ways to separate and assign the units are based on the spatiotemporal coordinates of the data. If there is an edge between two nodes in the graph, the corresponding hidden and visible subsets are connected in SRTRBM.

If the masks are fixed, the learning process of SRTRBM is the same as RTRBM. To learn the mask, the paper proposed to use the sigmoid function $\sigma(\nu_{ij})$ as a soft mask and learn the parameters $\nu_{ij}$. Also, a sparsity regularizer is added to $\nu_{ij}$s to encourage sparse connections. The training process is first to fix the mask while updating the other parameters and then update the mask. The author also gives the ssRBM version of SRTBM and RTRBM by replacing RBM with ssRBM while fixing the part of the energy function related to $\mathbf{r}_t$ to be unchanged. Experiments on human motion prediction, which is a TF-MPC task, and climate data prediction, which is an STSF-IG task, prove that SRTRBM achieves smaller one-step-ahead prediction error than RTRBM and RNN. This shows that using a masked weight matrix to model the spatial and temporal correlations is useful for STSF problems.

\textbf{Temporal Sigmoid Belief Network:}\quad
Gan et al.~\cite{gan2015deep} proposed a temporal extension of SBN called \emph{Temporal Sigmoid Belief Network} (TSBN). TSBN connects a sequence of SBNs in a way that the bias of the SBN at timestamp $t$ depends on the state of the previous SBNs. For a length-$T$ binary sequence $\mathbf{v}_{1:T}$, TSBN describes the following joint probability, in which we have omitted the biases for simplicity:
\begin{equation}
	\begin{aligned}
		&p_\theta(\mathbf{v}_{1:T}, \mathbf{h}_{1:T}) = p(\mathbf{h}_1) p(\mathbf{v}_1\mid \mathbf{h}_1) \\
    &\qquad\qquad\quad\prod_{t=2}^T p(\mathbf{h}_t \mid \mathbf{h}_{t-1}, \mathbf{v}_{t-1}) p(\mathbf{v}_t \mid \mathbf{h}_t, \mathbf{v}_{t-1}),\\
		&p(\mathbf{h}_{t, j} =1 \mid \mathbf{h}_{t-1}, \mathbf{v}_{t-1}) = \sigma(\mathbf{A}_j^T \mathbf{h}_{t-1}+ \mathbf{B}_j^T \mathbf{v}_{t-1}),\\
		&p(\mathbf{v}_{t, j} =1 \mid \mathbf{h}_{t}, \mathbf{v}_{t-1}) = \sigma(\mathbf{C}_j^T \mathbf{h}_{t} + \mathbf{D}_j^T \mathbf{v}_{t-1}).
	\end{aligned}
\end{equation}

Also, the author extended TSBN to the continuous domain by using a conditional Gaussian for the visible nodes:
\begin{equation}
	\begin{aligned}
		p(\mathbf{v}_t \mid  \mathbf{h}_t, \mathbf{v}_{t-1}) = \mathcal{N}(\bm{\mu}_t, \text{diag}(\bm{\sigma}_t^2)),\\
		\bm{\mu}_t = \mathbf{A}^T \mathbf{h}_{t-1} + \mathbf{B}^T \mathbf{v}_{t-1} + \mathbf{e},\\
		\log(\bm{\sigma}_t^2) = \mathbf{A^\prime}^T \mathbf{h}_{t-1} + \mathbf{B^\prime}^T \mathbf{v}_{t-1} + \mathbf{e^\prime}.
	\end{aligned}
\end{equation}

Multiple single-layer TSBNs can be stacked to form a deep architecture. For a multi-layer TSBN, the hidden states of the lower layers are conditioned on the states of the higher layers. The connection structure of a multi-layer TSBN is given in Figure~\ref{fig:dtgm-structures}(e). The author investigated both the stochastic and deterministic conditioning for multi-layer models and showed that the deep model with stochastic conditioning works better. Experiment results on the human motion prediction task show a multi-layer TSBN with stochastic conditioning outperforms variants of SRTRBM and RTRBM.

\subsection{FNN and RNN based Models}
One shortcoming of DTGMs for STSF is that the learning and inference processes are often complicated and time-consuming. Recently, there is a growing trend of using FNN and RNN based methods for STSF. Compared with DTGM, FNN and RNN based methods are simpler in the learning and inference processes. Moreover, these methods have achieved good performance in many STSF problems like video prediction~\cite{finn2016unsupervised, jia2016dynamic,wang2018predrnn}, precipitation nowcasting~\cite{xingjian2015convolutional,shi2017deep}, traffic speed forecasting~\cite{yu2017deep,li2018diffusion,zhang2018gaan}, human trajectory prediction~\cite{alahi2016social}, and human motion prediction~\cite{jain2016structural}. In the following, we will review these methods.

\subsubsection{Encoder-Forecaster Structure}
Recall that the goal of STSF is to predict the length-$L$ sequence in the future given the past observations. The \emph{Encoder-Forecaster} (EF) structure~\cite{nitish2015unsupervised,xingjian2015convolutional} first encodes the observations into a finite-dimensional vector or a probability distribution using the encoder and then generates the one-step-ahead prediction or multi-step predictions using the forecaster:
\begin{equation}
	\label{eq:ef-both-deterministic}
	\begin{aligned}
		\mathbf{s} &= f(\mathcal{F}_t;\theta_1),\\
		\hat{\mathbf{X}}_{t+1:t+L} &= g(\mathbf{s};\theta_2).
	\end{aligned}
\end{equation}
Here, $f$ is the encoder and $g$ is the forecaster. In our definition, we also allow probabilistic encoder and forecaster, where $\mathbf{s}$ and $\hat{\mathbf{X}}_{t+1:t+L}$ turn into random variables:
\begin{equation}
	\label{eq:ef-both-probabilstic}
	\begin{aligned}
		\mathbf{s} &\sim f(\mathcal{F}_t;\theta_1),\\
		\hat{\mathbf{X}}_{t+1:t+L} &\sim \pi_g(\mathbf{s};\theta_2).
	\end{aligned}
\end{equation}
Using probabilistic encoder and forecaster is a way to handle uncertainty and we will review the details in Section~\ref{ch:uncertainty}. All of the surveyed FNN and RNN based methods fit into the EF framework. In the following, we will review how previous works design the encoder and the forecaster for different types of STSF problems, i.e., STSF-RG, STSF-IG, and TF-MPC.
\subsubsection{Models for STSF-RG}
\label{ch:model-stsf-rg}
For STSF-RG problems, the spatiotemporal sequences lie in a regular grid and can naturally be modeled by CNNs, which have proved effective for extracting features from images and videos~\cite{alex2012imagenet,tran2015learning}. Therefore, researchers have used CNNs to build the encoder and the forecaster. In~\cite{goroshin2015learning}, two input frames are encoded independently to two state vectors by a 2D CNN. The future frame is generated by linearly extrapolating these two state vectors and then mapping the extrapolated vector back to the image space with a series of 2D deconvolution and unpooling layers. Also, to better capture the spatial features, the author proposed the phase pooling layer, which keeps both the pooled values and pooled indices in the feature maps. In~\cite{mathieu2016deep}, multi-scale 2D CNNs with ReLU activations were used as the encoder and the forecaster. Rather than encoding the frames independently, the author concatenated the input frames along the channel dimension to generate a single input tensor. The input tensor is rescaled to multiple resolutions and encoded by a multi-scale 2D CNN. To better capture the spatiotemporal correlations, Vondrick et al.~\cite{vondrick2016generating} proposed to use 3D CNNs as the encoder and forecaster. Also, the author proposed to use a 2D CNN to predict the background image and only use the 3D CNN to generate the foreground. Later, Shi et al.~\cite{shi2017deep} compared the performance of 2D CNN and 3D CNN for precipitation nowcasting and showed that 3D CNN outperformed 2D CNN in the experiments. Kalchbrenner et al.~\cite{kalchbrenner2016video} proposed the \emph{Video Pixel Network} (VPN) that uses multiple causal convolution layers~\cite{DBLP:journals/corr/OordDZSVGKSK16} in the forecaster. Unlike the previous 2D CNN and 3D CNN models, of which the forecaster directly generates the multi-step predictions, VPN generates the elements in the spatiotemporal sequence one-by-one with the help of causal convolution. Experiments show that VPN outperforms the baseline which uses a 2D CNN to generate the predictions directly. Recently, Xu et al.~\cite{xu2018predcnn} proposed the PredCNN network that stacks multiple dilated causal convolution layers to encode the input frames and predict the future. Experiments show that PredCNN achieves state-of-the-art performance in video prediction.

\begin{figure}[tb!]
  \centering 
  \subfigure[Structure of ConvRNN.]{
    \includegraphics[width=0.35\textwidth]{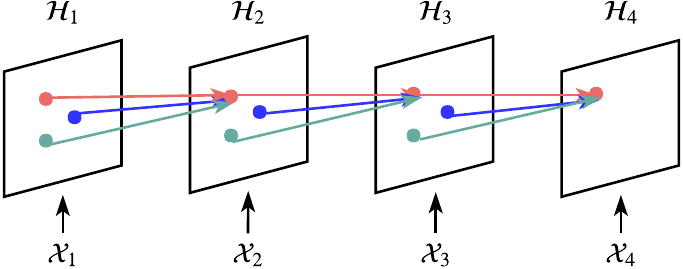}
    \label{fig:conv_rnn}
  }
  ~
  \subfigure[Structure of TrajRNN.]{
    \includegraphics[width=0.35\textwidth]{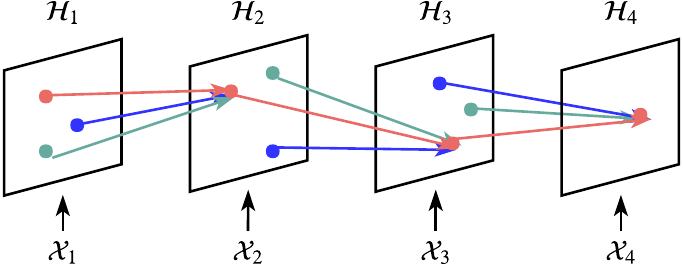}
    \label{fig:traj_rnn}
  }
  \caption{Illustration of the connection structures of ConvRNN, TrajRNN. Links with the same color share the same transition weights. Source:~\cite{shi2017deep}.}
  \label{fig:conv_rnn_traj_rnn}
\end{figure}

\begin{figure}[tb!]
  \centering 
  \subfigure[Structure of a stack of three ConvLSTM layers.]{
    \includegraphics[width=0.35\textwidth]{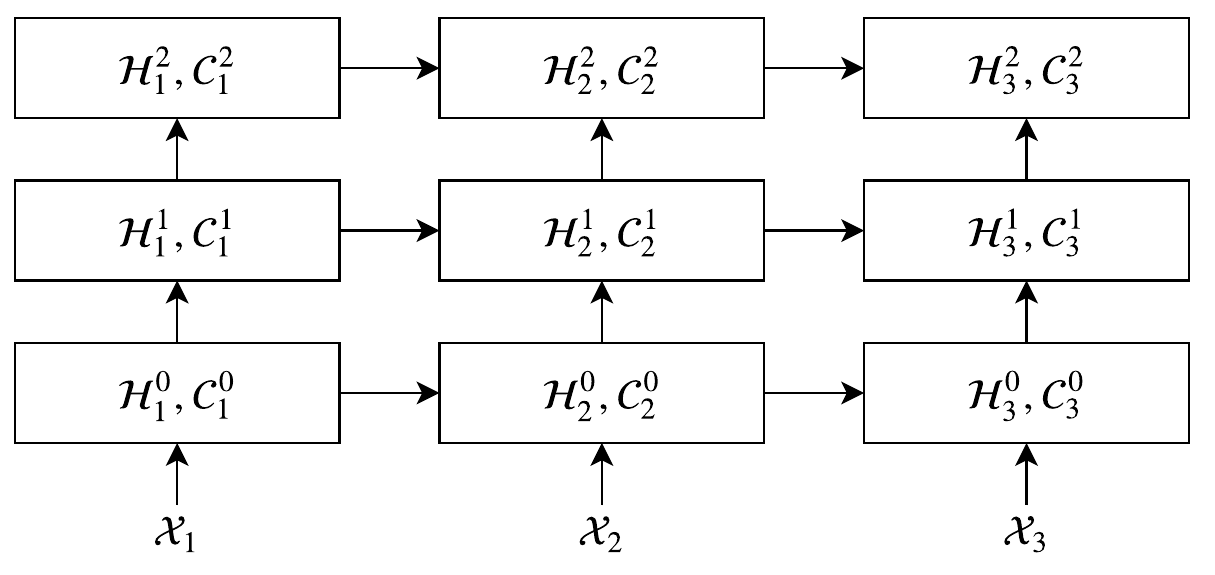}
    \label{fig:stack_convlstm}
  }
  ~
  \subfigure[Structure of PredRNN with three layers.]{
    \includegraphics[width=0.35\textwidth]{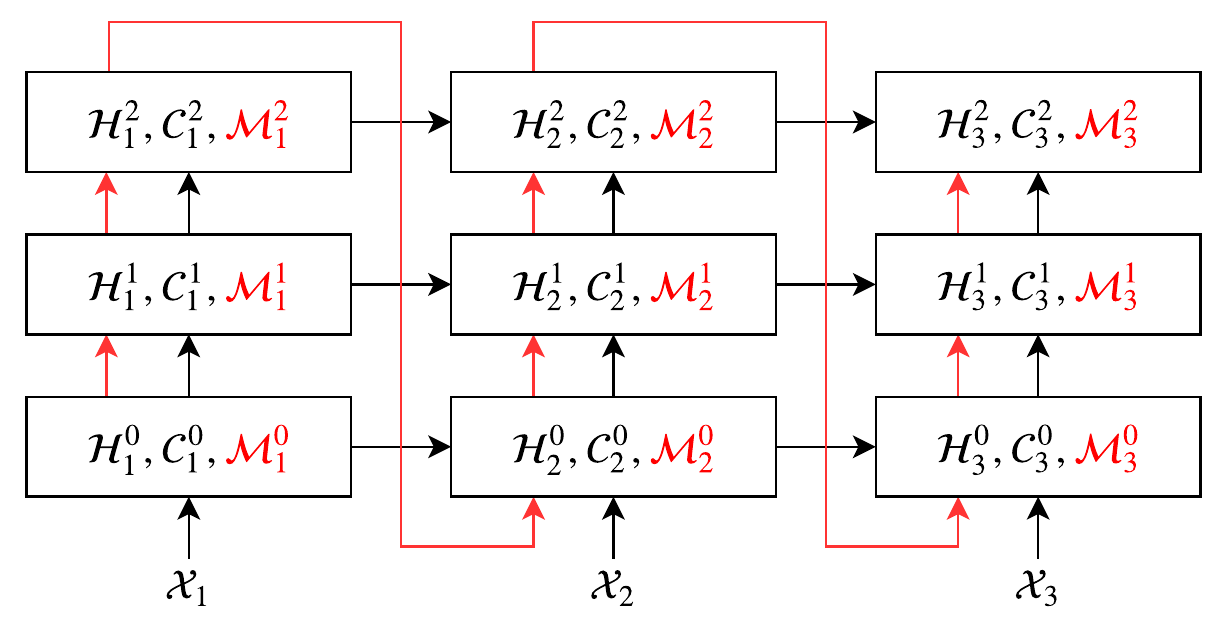}
    \label{fig:predrnn}
  }
  \caption{Comparison of multi-layer ConvLSTMs and the PredRNN. $\mathcal{M}_t$ is the global spatiotemporal memory that is updated along the red links.}
  \label{fig:stack_conv_rnn_pred_rnn}
\end{figure}

Besides using CNNs, researchers have also used RNNs to build models for STSF-RG. The advantage of using RNN for STSF problems is that it naturally models the temporal correlations within the data. Srivastava et al.~\cite{nitish2015unsupervised} proposed to use multi-layer LSTM networks as the encoder and the forecaster. To use the LSTM, the author directly flattened the images into vectors. Oh et al.~\cite{oh2015action} proposed to use 2D CNNs to encode the image frames before feeding into LSTM layers. However, the state-state connection structure of \emph{Fully-Connected LSTM} (FC-LSTM) is redundant for STSF and is not suitable for capturing the spatiotemporal correlations in the data~\cite{xingjian2015convolutional}. To solve the problem, Shi et al.~\cite{xingjian2015convolutional} proposed the \emph{Convolutional LSTM} (ConvLSTM). ConvLSTM uses convolution instead of full-connection in both the feed-forward and recurrent connections of LSTM. The formula of ConvLSTM is given in the following:
\begin{equation}
  \label{eq:convlstm}
  \begin{aligned}
    \mathcal{I}_t &= \sigma(\mathcal{W}_{xi} \ast \mathcal{X}_t + \mathcal{W}_{hi} \ast \mathcal{H}_{t-1} + b_i),\\
    \mathcal{F}_t &= \sigma(\mathcal{W}_{xf} \ast \mathcal{X}_t + \mathcal{W}_{hf} \ast \mathcal{H}_{t-1} + b_f ),\\
    \mathcal{C}_t &= \mathcal{F}_t \circ \mathcal{C}_{t-1} +\\
    &\qquad\mathcal{I}_t \circ tanh(\mathcal{W}_{xc} \ast X_t + W_{hc} \ast \mathcal{H}_{t-1} + b_c),\\
    \mathcal{O}_t &= \sigma(\mathcal{W}_{xo} \ast \mathcal{X}_t + \mathcal{W}_{ho} \ast \mathcal{H}_{t-1} + b_o),\\
    \mathcal{H}_t &= \mathcal{O}_t \circ tanh(\mathcal{C}_t).
  \end{aligned}
\end{equation}
Here, all the states, cells, and gates are three-dimensional tensors. The author used two ConvLSTM layers as the encoder and the forecaster and showed that 1) ConvLSTM outperforms FC-LSTM when applied to the precipitation nowcasting problem and 2) the two-layer model performs better than the single-layer model. Similar to the way of extending LSTM to ConvLSTM, other types of RNNs, like GRU, can be extended to \emph{Convolutional RNNs} (ConvRNNs) by using convolution in state-state transitions. Since ConvRNN combines the advantage of CNN and RNN, it is widely adopted in later researches~\cite{finn2016unsupervised,jia2016dynamic,ruben2017decomposing,apratim18cvprb,jang2018video} as a basic block for building DL models for STSF-RG. There are also attempts to improve the structure of ConvRNN. Shi et al.~\cite{shi2017deep} proposed the \emph{Trajectory GRU} (TrajGRU) model that improves ConvGRU by actively learning the recurrent connection structure. Because the convolutional recurrence structure adopted in ConvRNNs is location-invariant, it is not suitable for modeling location-variant motions, e.g., rotation and scaling. Thus, TrajGRU uses a learned subnetwork to output the recurrent connection structures:
\begin{equation}
    \begin{aligned}
        \mathcal{U}_t, \mathcal{V}_t &= \gamma(\mathcal{X}_t, \mathcal{H}_{t-1}),\\
        \mathcal{Z}_t &= \sigma(\mathcal{W}_{xz} \ast \mathcal{X}_t + \sum_{l=1}^L{\mathcal{W}^l_{hz} \ast \mathcal{H}_{t-1, \mathcal{U}_{t, l}, \mathcal{V}_{t, l}}}),\\
        \mathcal{R}_t &= \sigma(\mathcal{W}_{xr} \ast \mathcal{X}_t + \sum_{l=1}^L{\mathcal{W}^l_{hr} \ast \mathcal{H}_{t-1, \mathcal{U}_{t, l}, \mathcal{V}_{t, l}}}),\\
        \mathcal{H}^\prime_t &= f(\mathcal{W}_{xh} \ast \mathcal{X}_t +\\
         &\qquad\mathcal{R}_{t} \circ (\sum_{l=1}^L{\mathcal{W}^l_{hh} \ast \mathcal{H}_{t-1, \mathcal{U}_{t, l}, \mathcal{V}_{t, l}})}),\\
        \mathcal{H}_t &= (1 - \mathcal{Z}_t) \circ \mathcal{H}^\prime_t + \mathcal{Z}_t \circ \mathcal{H}_{t-1}.
    \end{aligned}
\end{equation}
Here, $\gamma$ is a convolutional subnetwork that outputs $L$ couples of flow fields $\mathcal{U}_t, \mathcal{V}_t \in \mathbb{R}^{L \times H \times W}$. $\mathcal{H}_{t-1, \mathcal{U}_{t, l}, \mathcal{V}_{t,l}}$ means to shift the elements in $\mathcal{H}_{t-1}$ through the flow field $\mathcal{U}_{t,l}, \mathcal{V}_{t,l}$ by bilinear warping~\cite{jaderberg2015spatial}. The connection structures of ConvRNN and TrajRNN are illustrated in Figure~\ref{fig:conv_rnn_traj_rnn}. Experiments on both the synthetic dataset and the real-world HKO-7 benchmark show that TrajGRU outperforms 2D CNN, 3D CNN, and ConvGRU. Wang et al.~\cite{wang2017predrnn} proposed the \emph{Predictive RNN} (PredRNN), which extends ConvLSTM by adding a new type of spatiotemporal memory. Unlike ConvLSTM, of which the memory states are constrained inside each LSTM layer, PredRNN maintains another global memory state to store the spatiotemporal information in the lower layers and in the previous timestamps. In PredRNN, memory states are allowed to zigzag across the RNN layers. The structure of PredRNN is illustrated in Figure~\ref{fig:predrnn}. Here, $\mathcal{M}^l_t$s are the global spatiotemporal memory states and are updated in a similar way as the cell states in ConvLSTM except that it is updated in a zigzag order. Based on PredRNN, Wang et al.~\cite{wang2018predrnn} proposed the PredRNN++ structure which added more nonlinearities when updating $\mathcal{M}^l_t$s. Experiments show that PredRNN++ outperforms both PredRNN and other baseline models including TrajGRU and variants of ConvLSTM.

Apart from studying the basic building blocks like CNNs and RNNs, another line of research tries to disentangle the motion and content of the spatiotemporal sequence to better forecast the future. Jia et al.~\cite{jia2016dynamic} and Finn et al.~\cite{finn2016unsupervised} proposed to predict how the frames will transform instead of directly predicting the future frames. Finn et al.~\cite{finn2016unsupervised} proposed two transformation methods named \emph{Convolutional Dynamic Neural Advection} (CDNA) and \emph{Spatial Transformer Predictors} (STP). To predict the frame at timestamp $t+1$, CDNA generates $N$ 2D convolutional filters $\{\mathbf{W}_t^1, \mathbf{W}_t^2, ..., \mathbf{W}_t^N\}$ and a masking tensor $\mathcal{M} \in \mathbb{R}^{N \times H \times W}$ where $\sum_{c=1}^N \mathcal{M}_{c, i, j} = 1$. The previous frame $\mathcal{X}_{t}$ is transformed $N$ times by these filters and these $N$ transformed frames are linearly combined by the masking tensor to generate the final prediction. STP is similar to CDNA except that the $\mathbf{W}_t^n$s become the parameters of different affine transformations~\cite{jaderberg2015spatial}.
Similar to CDNA and STP, Jia et al.~\cite{jia2016dynamic} proposed the \emph{Dynamic Filter Network} (DFN) that directly outputs the parameters of a $K \times K$ local filter for every location. The prediction is generated by applying these filters to the previous frame. Actually, DFN can be viewed as a special case of CDNA. If we manually fix the 2-dimensional convolutional filters in CDNA to have $K^2$ elements, of which only a single element is non-zero, CDNA turns into DFN. Villegas et al.~\cite{ruben2017decomposing} proposed \emph{Motion-Content Network} (MCnet) which uses two separated networks to encode the motion and the content. In MCnet, the input of the motion encoder is the sequence of difference images $\{\mathcal{X}_{2} - \mathcal{X}_1, ..., \mathcal{X}_t - \mathcal{X}_{t-1}\}$ and the input of the content encoder is the last observed image $\mathcal{X}_t$. The encoded motion features and content features are fused together to generate the prediction. Experiments show that disentangling the motion and the content helps improve the performance. This architecture has been used in other papers~\cite{xu2018structure,jang2018video} to enhance the prediction performance.


\subsubsection{Models for STSF-IG}
\label{ch:model-stsf-ig}

Unlike STSF-RG problems, the measurements in STSF-IG problems are observed in some sparsely distributed stations which are more challenging to deal with. The simple strategy is to model the measurements observed at each station independently, which has been adopted in Yu et al.~\cite{yu2017deep}. However, this strategy has not considered the correlation among different stations. A recent trend in solving this problem is to convert these sparsely distributed stations to a graph and utilize graph convolution to build the encoder and the forecaster. We will mainly introduce these graph convolution-based methods in this section.

Li et al.~\cite{li2018diffusion} proposed the \emph{Diffusion Convolutional Recurrent Neural Network} (DCRNN) for traffic speed forecasting. To convert the traffic stations into a graph, the author generated the adjacency matrix based on the thresholded pairwise road network distances. After that, DCRNN is used to encode the sequence of graphical data and predict the future. The formula of DCRNN is similar to ConvRNN except that the \emph{Diffusion Convolution} (DC) is used in the input-state and state-state transitions. DC is a type of graph convolution that considers the direction information in the graph, which has the following formula:
\begin{equation}
\label{eq:dcrnn}
\begin{aligned}
\mathbf{Y} = \sum_{k=0}^{K - 1} (\theta_{k, 1} (\mathbf{D}_O^{-1} \mathbf{A})^k) + \theta_{k, 2}(\mathbf{D}_I^{-1} \mathbf{A}^T)^k) \mathbf{X},
\end{aligned}
\end{equation}
where $\mathbf{X} \in \mathbb{R}^{N \times C}$ is the input features, $\mathbf{A}$ is the adjacency matrix, $\mathbf{D}_O$ is the diagonal out-degree matrix, $\mathbf{D}_I$ is the diagonal in-degree matrix, $\theta_{k,1}, \theta_{k,2}$ are the parameters, $K$ is the number of diffusion steps, $N$ is the number of nodes in the graph, and $C$ is the number of the features. Unlike other graph convolution operators like ChebNet~\cite{defferrard2016convolutional} which operates on undirected graphs, DC operates on a directed graph and is more suitable for real-life scenarios. Experiments on two real-world traffic datasets show that DCRNN outperforms FC-LSTM and the ChebNet-based \emph{Graph Convolutional Recurrent Neural Network} (GCRNN)~\cite{li2018diffusion}. Later, Zhang et al.~\cite{zhang2018gaan} proposed the \emph{Graph GRU} (GGRU) network that generalizes the idea of DCRNN. Instead of proposing a single model, the author proposed a unified method for constructing an RNN based on an arbitrary graph convolution operator. The formula of GGRU is given in the following:
\begin{equation}
  \label{eq:graph_gru}
  \begin{aligned}
      \mathbf{U}_t = &\sigma(\Gamma_{\Theta_{xu}}(\mathbf{X}_t, \mathbf{X}_t; \mathcal{G}) \\
      &\quad + \Gamma_{\Theta_{hu}}(\mathbf{X}_t \oplus \mathbf{H}_{t-1}, \mathbf{H}_{t-1}; \mathcal{G})),\\
      \mathbf{R}_t = &\sigma(\Gamma_{\Theta_{xr}}(\mathbf{X}_t, \mathbf{X}_t; \mathcal{G})\\
      &\quad + \Gamma_{\Theta_{hr}}(\mathbf{X}_t \oplus \mathbf{H}_{t-1}, \mathbf{H}_{t-1}; \mathcal{G})),\\
      \mathbf{H}^\prime_t =& h(\Gamma_{\Theta_{xh}}(\mathbf{X}_t, \mathbf{X}_t; \mathcal{G})\\
      &\quad + \mathbf{R}_t \circ \Gamma_{\Theta_{hh}}(\mathbf{X}_t \oplus \mathbf{H}_{t-1}, \mathbf{H}_{t-1}; \mathcal{G})),\\
      \mathbf{H}_t =& (1 - \mathbf{U}_t) \circ \mathbf{H}^\prime_t + \mathbf{U}_t \circ \mathbf{H}_{t-1}.
  \end{aligned}
\end{equation}
Here, $\mathbf{U}_t, \mathbf{R}_t$ are the update gate and reset gate that controls the memory flow, $\mathbf{H}_t$ is the hidden state, $\mathbf{X}_t$ is the input, $h(\cdot)$ is the activation function, and $\Gamma(\cdot, \cdot)$ is an arbitrary graph convolution operator. The author also proposed a new multi-head gated graph attention aggregator that uses neural attention to aggregate the neighboring features. Experiments show that GGRU outperforms DCRNN if the gated attention aggregator is used.
Yu et al.~\cite{yu2018spatio} proposed the \emph{Spatio-Temporal Graph Convolutional Network} (STGCN) that applies graph convolution in a different manner. The author directly stacked multiple \emph{Spatio-Temporal Convolution} (ST-Conv) blocks to build the network. The ST-Conv block is a concatenation of two temporal convolution layers and one graph convolution layer.  For the graph convolution operator, the paper tested the ChebNet and its first-order approximation~\cite{kipf2017semi} and found that ChebNet performed better. The major advantage of STGCN over RNN-based methods is it is faster in the training phase.

\subsubsection{Models for TF-MPC}
\label{ch:model-TF-MPC}
The goal of TF-MPC is to predict the future trajectories of all objects in the moving point cloud. Fragkiadaki et al.~\cite{fragkiadaki2015recurrent} proposed a basic RNN based method for this problem. In the model, the coordinate sequence of each human joint is encoded by the same LSTM. The drawback of this basic method is that it does not model the correlation among the individuals. To solve the problem, Alahi et al. proposed the SocialLSTM model~\cite{alahi2016social} that uses a set of interacted LSTMs to model the trajectory of the crowd. In SocialLSTM, the position of each person is modeled by an LSTM and a social pooling layer interconnects these LSTMs. Social pooling aggregates the hidden states of the nearby LSTMs at the previous timestamp and use this aggregated state to help decide the current state. The transition rule of the SocialLSTM is given in the following:
\begin{equation}
  \label{eq:social-lstm}
  \begin{aligned}
    \mathcal{N}^i_t &= \{j \mid j\in \mathcal{N}_i, \abs{x_t^j - x_t^i} \leq m, \abs{y_t^j - y_t^i} \leq n\},\\
    \mathbf{H}_t^i &= \sum_{j\in \mathcal{N}^i_t} \mathbf{h}_{t-1}^j,\\
    \mathbf{e}_t^i &= \phi_1(x_t^i, y_t^i; \mathbf{W}_e),\\
    \mathbf{a}_t^i &= \phi_2(\mathbf{H}_t^i; \mathbf{W}_l),\\
    \mathbf{h}_t^i, \mathbf{c}_t^i &= \text{LSTM}(\mathbf{h}_{t-1}^i, \mathbf{c}_{t-1}^i, \text{vec}(\mathbf{e}_t^i, \mathbf{a}_t^i); \mathbf{W}_l).
  \end{aligned}
\end{equation}
Here, $\mathbf{H}_t^i$ is the aggregated state vector in the social pooling step, $(x_t^j, y_t^j)$ is the coordinate of the $j$th person at timestamp $t$, $\phi_1, \phi_2$ are mapping functions, and $m, n$ are distance thresholds. Experiments show that adding the social pooling layer is helpful for the TF-MPC task. In~\cite{irtiza2018mx}, the author proposed an improved version of social pooling called \emph{View Frustum Of Attention} (VFOA) social pooling. This method defines the neighborhood set in the social pooling layer based on the calculated VFOA interest region. Jain et al.~\cite{jain2016structural} proposed the \emph{Structural-RNN} (S-RNN) network to solve the human motion prediction problem. S-RNN imposes structures in the recurrent connections based on a given spatiotemporal graph. For human motion prediction, this spatiotemporal graph is constructed by the relationship between different human joints. S-RNN maintains two types of RNNs, nodeRNN and edgeRNN. The outputs of edgeRNNs are fed to the corresponding nodeRNNs.
Experiments show that the S-RNN achieves the state-of-the-art result in the human motion prediction problem.

\subsubsection{Methods for Handling Uncertainty}
\label{ch:uncertainty}
Most real-world dynamical systems, e.g., atmosphere, are inherently stochastic and unpredictable. Simply assuming the outcome is deterministic, which is common in FNN and RNN based models, will generate blurry predictions. Thus, recent studies~\cite{goroshin2015learning,vondrick2016generating,babaeizadeh2018stochastic} have proposed ways for handling uncertainty.

In~\cite{fragkiadaki2015recurrent}, the author proposed to use a probabilistic forecaster that outputs the parameters of a \emph{Gaussian Mixture Model} (GMM). This enables the network to generate stochastic predictions. The author found that the smallest forecasting error was always produced by the most probable sample, which was similar to the output of a model trained by the Euclidean loss. Apart from using a probabilistic forecaster, the author also proposed a curriculum learning strategy which gradually adds noise to the input during training. This can be viewed as regularizing the model not to be overconfident about its predictions. This technique has also been adopted in later works~\cite{jain2016structural}. Goroshin et al.~\cite{goroshin2015learning} proposed to add a latent variable $\delta$ in the network to represent the uncertainty of the prediction. The latent variable $\delta_i$ is not determined by the input $\mathcal{X}_i$ and can be adjusted in the learning process to minimize the loss. Before updating the parameters of the network, the algorithm first updates $\delta_i$ for $k$ steps where $k$ is a hyperparameter. At test time, the $\delta_i$s is sampled based on its distribution on the training set.

Another approach to handle uncertainty is to utilize the recently popularized conditional \emph{Generative Adversarial Network} (GAN)~\cite{goodfellow2014generative,DBLP:journals/corr/MirzaO14}. Mathieu et al.~\cite{mathieu2016deep} proposed to use conditional GAN as the loss function to train the multi-scale 2D CNN. The GAN loss used in the paper is given as follows:
\begin{equation}
\label{eq:gan}
\begin{aligned}
&\min_{\theta_G} \max_{\theta_D} \mathbb{E}_{\mathcal{Y} \sim p_{\text{data}}(\mathcal{Y} \mid \mathcal{X})}[\log D_{\theta_D}(\mathcal{X}, \mathcal{Y})]\\
&\quad+\mathbb{E}_{z \sim p_z} [\log(1 - D_{\theta_D}(\mathcal{X}, G_{\theta_G}(\mathcal{X}, z)))].
\end{aligned}
\end{equation}
Here, $D_{\theta_D}$ is the discriminator, $G_{\theta_G}$ is the forecasting model, $\mathcal{X}$ is the observed sequence, $\mathcal{Y}$ is the ground-truth sequence, and $p(z)$ is the noise distribution. The discriminator is trained to differentiate between the ground-truth output and the prediction generated by the forecaster. On the other hand, the forecasting model is trained to fool the discriminator. The final loss function in the paper combined the GAN loss, the L2 loss, and the image gradient difference loss. Experiments show that the model trained with GAN generates sharper predictions. Vondrick et al.~\cite{vondrick2016generating} used a similar GAN loss to train the 3D CNN. Jang et al.~\cite{jang2018video} proposed to use two discriminators to respectively discriminate the motion and appearance part of the generated samples.

Besides GAN, the \emph{Variational Auto-encoder} (VAE)~\cite{kingma2014auto} has also been adopted for dealing with the uncertainty in STSF. Babaeizadeh et al.~\cite{babaeizadeh2018stochastic} proposed \emph{Stochastic Variational Video Prediction} (SV2P) which uses the following objective function:
\begin{equation}
\label{eq:sv2p}
\begin{aligned}
l &= - \mathbb{E}_{q_\phi(\mathbf{Z} \mid \mathbf{X}_{1:T})}[\log p_\theta(\mathbf{X}_{t:T}\mid \mathbf{X}_{1:t}, \mathbf{Z})] \\
&\quad\quad+ D_{\text{KL}}(q_\phi(\mathbf{Z} \mid \mathbf{X}_{1:T}) || p(\mathbf{Z})),
\end{aligned}
\end{equation}
where $q_\phi$ is the variational inference network that approximates the posterior of the latent variable $\mathbf{Z}$ and $p_\theta$ is an RNN generator. The experiments show that adding VAE greatly alleviates the blurry problem of the predictions. Later, Denton \& Fergus~\cite{denton2018stochastic} proposed the \emph{Stochastic Video Generator with Learned Prior} (SVG-LP) which uses a learned time-varying prior $p_\psi(\mathbf{Z}_t \mid \mathbf{X}_{1:t-1})$ to replace the fixed prior $p(\mathbf{Z})$ in~\eqref{eq:sv2p}. The extension in their paper is similar to the \emph{Variational RNN} (VRNN)~\cite{chung2015a}. Learning a time-variant prior can be interpreted as learning a predictive model of uncertainty. The author showed in the experiments that using a learned prior improves the forecasting performance.

\begin{table*}[tb!]
\centering
\caption{Summary of the reviewed methods for STSF. We put ``$\surd$'' under STSF-RG, STSF-IG, and TF-MPC to show the method has been extended to solve the problem. We put ``$\surd$'' under ``Unc.'' to show some of the methods have handled uncertainty.}
\label{tbl:overall-summary}
\begin{tabular}{|l|l|l|c|c|c|c|}
\hline
Category & Subcategory & Methods & STSF-RG & STSF-IG & TF-MPC & Unc. \\ \hline
\multirow{6}{*}{Classical} & \multirow{2}{*}{Feature-based} & Spatiotemporal indicator~\cite{ohashi2012wind} &  & $\surd$ &  &  \\ \cline{3-7} 
 &  & Multiple predictors~\cite{zheng2015forecasting} &  & $\surd$ &  &  \\ \cline{2-7} 
 & \multirow{3}{*}{SSM} & Group-based~\cite{asahara2011pedestrian,mathew2012predicting} &  &  & $\surd$ & $\surd$ \\ \cline{3-7} 
 &  & STARIMA~\cite{cliff1975model, cliff1975space, pfeifer1980starima} & $\surd$ & $\surd$ &  & $\surd$ \\ \cline{3-7} 
 &  & Low-rankness~\cite{bahadori2014fast, yu2015accelerated, yu2016learning} & $\surd$ & $\surd$ &  & $\surd$ \\ \cline{2-7} 
 & GP & GP~\cite{trautman2010unfreezing,flaxman2015fast,flaxman2015machine,senanayake2016predicting} & $\surd$ & $\surd$ & $\surd$ & $\surd$ \\ \hline
\multirow{17}{*}{DL} & \multirow{4}{*}{DTGM} & TRBM~\cite{sutskever2007learning} & $\surd$ &  & $\surd$ & $\surd$ \\ \cline{3-7} 
 &  & RTRBM~\cite{sutskever2009recurrent} & $\surd$ &  & $\surd$ & $\surd$ \\ \cline{3-7} 
 &  & SRTRBM~\cite{mittelman2014structured} & $\surd$ & $\surd$ & $\surd$ & $\surd$ \\ \cline{3-7} 
 &  & TSBN~\cite{gan2015deep} & $\surd$ &  & $\surd$ & $\surd$ \\ \cline{2-7} 
 & \multirow{13}{*}{FNN \& RNN} & FC-LSTM & $\surd$ & $\surd$ & $\surd$ & $\surd$ \\ \cline{3-7} 
 &  & 2D CNN~\cite{oh2015action,mathieu2016deep} \& 3D CNN~\cite{vondrick2016generating} & $\surd$ &  &  & $\surd$ \\ \cline{3-7} 
 &  & ConvLSTM~\cite{xingjian2015convolutional} \& TrajGRU~\cite{shi2017deep} & $\surd$ &  &  &  \\ \cline{3-7} 
 &  & PredRNN~\cite{wang2017predrnn,wang2018predrnn} & $\surd$ &  &  &  \\ \cline{3-7} 
 &  & VPN~\cite{kalchbrenner2016video} \& PredCNN~\cite{xu2018predcnn} & $\surd$ &  &  &  \\ \cline{3-7} 
 &  & Learn transformation~\cite{finn2016unsupervised,jia2016dynamic} & $\surd$ &  &  & \\ \cline{3-7} 
 &  & Motion + Content~\cite{ruben2017decomposing,xu2018structure,jang2018video} & $\surd$ &  &  & $\surd$ \\ \cline{3-7} 
 &  & DCRNN~\cite{li2018diffusion} \& GGRU~\cite{zhang2018gaan} &  & $\surd$ &  &  \\ \cline{3-7} 
 &  & STGCN~\cite{yu2018spatio} &  & $\surd$ &  &  \\ \cline{3-7} 
 &  & Social pooling~\cite{alahi2016social,irtiza2018mx} &  &  & $\surd$ & $\surd$ \\ \cline{3-7} 
 &  & S-RNN~\cite{jain2016structural} &  &  & $\surd$ & $\surd$ \\ \cline{3-7} 
 &  & $\delta$-gradient~\cite{goroshin2015learning} & $\surd$ &  &  & $\surd$ \\ \cline{3-7} 
 &  & SV2P~\cite{babaeizadeh2018stochastic} \& SVG-LP~\cite{denton2018stochastic} & $\surd$ &  &  & $\surd$ \\ \hline
\end{tabular}
\end{table*}

\subsection{Remarks}
\label{ch:deep-remark}
In this section, we reviewed two types of DL based methods for STSF: DTGMs and FNN and RNN based methods. The majority of the reviewed DTGMs treat STSF as a general multivariate sequence forecasting problem and SRTRBM is the only model that gives special treatment of the spatiotemporal sequences. Also, learning a DTGM requires approximation and sampling techniques and is more complicated compared with the FNN and RNN based methods. However, DTGM can well capture the uncertainty in the data due to its stochastic generative process. On the other hand, the FNN and RNN based models are easy to train and fast for prediction. We reviewed the representative methods for STSF-RG, STSF-IG, and TF-MPC. Although these three types of STSF problems involve different tasks, the proposed methods for these problems have strong relationships. For example, ConvLSTM, DCRNN, and SocialLSTM, which are respectively proposed for STSF-RG, STSF-IG, and TF-MPC, improve upon RNN by introducing structured recurrent connections. To be more specific, ConvLSTM uses convolution in recurrent transitions, DCRNN uses the graph convolution, and SocialLSTM uses the social pooling layer. Therefore, the success of these models greatly relies on the appropriate design of the basic network architecture. Also, FNN and RNN based methods are usually deterministic and are not well-designed for capturing the uncertainty in the data. We thus reviewed some recently popularized techniques for handling uncertainty including GAN and VAE, which have improved the forecasting performance.

\section{Summary and Future Works}
\label{ch:discussion}

In this survey, we reviewed the machine learning based methods for STSF. We first examined the general strategies for multi-step sequence forecasting, which contain the IMS, DMS, boosting strategy, and scheduled sampling. We showed that IMS and DMS have their advantages and disadvantages. The boosting strategy and scheduled sampling can be viewed as the mid-grounds between these two approaches. Next, we reviewed the specific methods for STSF. We divided these methods into two major categories: classical methods and DL based methods. As mentioned in~\ref{ch:classical-remark} and~\ref{ch:deep-remark}, the existing classical methods and deep learning methods have their own trade-offs when solving the STSF problem. The overall summarization of the reviewed methods is given in Table~\ref{tbl:overall-summary}.

There are many future research directions for machine learning for STSF. 1) We can enhance the strategies for the multi-step sequence forecasting. One direction is to design new training curriculum to shift from IMS to DMS. For example, we can extend the forecasting horizon exponentially from $1$ to $N$ to better model longer sequences, which is common in TF-MPC problems. Also, we can generalize the boosting method described in Section~\ref{ch:boosting} to the general STSF problems. In addition, we can reduce the complexity of DMS from $O(h)$ to $O(\log h)$ by training $O(\log h)$ models that generate the predictions of horizons $2^0, 2^1,...,2^{\log h-1}$. We can use these models to make prediction for any horizon up to $h$ in the spirit of binary coding. Moreover, the objective function mentioned in Section~\ref{ch:scheduled-sampling} will not be differentiable due to the sampling process and the original paper has just ignored this problem. This problem can be solved by approximating the gradient through techniques in reinforcement learning~\cite{sutton1999policy}. 2) We can propose new FNN and RNN based methods for STSF-IG problems. Compared with STSF-RG, DL for STSF-IG is less mature and we can bring the ideas in STSF-RG, like decomposing motion and content, adding a global memory structure, and using GAN/VAE to handle uncertainty, to STSF-IG. 3) We can combine DTGMs with FNN and RNN based methods to get the best of both worlds. The generation flavor of DTGMs has its role to play in FNN and RNN based models particularly for handling uncertainty. One feasible approach is to use the idea of \emph{Bayesian Deep Learning} (BDL)~\cite{Wang:2016:TBD:3024719.3024768}, in which FNN and RNN based models are used for perception and DTGMs are used for inference. 4) We can study the type of TF-MPC problem where both the measurements and coordinates are changing. We can consider to solve the ``video object prediction'' problem where we have the bounding boxes of the objects at the previous frames and the task is to predict both the objects' future appearances and bounding boxes. 5) We can study new learning scenarios like online learning and transfer learning for STSF. For the STSF problem, the online learning scenario is essential because spatiotemporal data, like regional rainfall, usually arrive in an online fashion~\cite{shi2017deep}. The transfer learning scenario is also essential because STSF can act as a pretraining method. The model trained by STSF can be transferred to solve other tasks like classification and control~\cite{finn2016unsupervised,finn2017deep}. Designing a better transfer learning method for STSF will improve the performance of these tasks. 


%





\ifCLASSOPTIONcaptionsoff
  \newpage
\fi



\bibliographystyle{IEEEtran}
\bibliography{general,irregular,regular,trajectory,other,multistep,uncertainty}

%

\begin{IEEEbiography}{Xingjian Shi}
received his BEng degree in information security from Shanghai Jiao Tong University, China. He is currently a Ph.D student at the Department of Computer Science and Engineering of Hong Kong University of Science and Technology. His research interests are in deep learning and spatiotemporal analysis. He received the Hong Kong PhD Fellowship in 2014.
\end{IEEEbiography}

\begin{IEEEbiography}{Dit-Yan Yeung}
received his BEng degree in electrical engineering and MPhil degree in computer science from the University of Hong Kong, and PhD degree in computer science from the University of Southern California. He started his academic career as an assistant professor at the Illinois Institute of Technology in Chicago. He then joined the Hong Kong University of Science and Technology where he is now a full professor and the Acting Head of the Department of Computer Science and Engineering. His research interests are in computational and statistical approaches to machine learning and artificial intelligence.
\end{IEEEbiography}

\vfill



\end{document}